%% file: main.tex
\documentclass[runningheads]{llncs}
\usepackage{eccv}
\input{preamble}

\title{\dataname: A Multiview Synthetic 4D Dataset} 
\titlerunning{\dataname}
\author{%
Zeren Jiang\textsuperscript{*}\inst{1} \and
Yushi Lan\textsuperscript{*}\inst{1} \and
Yihang Luo\inst{2} \and
Yufan Deng\inst{1} \and
Zihang Lai\inst{1} \and
\\
Edgar Sucar\inst{1} \and
Christian Rupprecht\inst{1} \and
Iro Laina\inst{1} \and
Diane Larlus\inst{3} \and
\\
Chuanxia Zheng\inst{2} \and
Andrea Vedaldi\inst{1}
}
\authorrunning{Z.~Jiang et al.}
\institute{VGG, University of Oxford \and
Nanyang Technological University \and
Naver Labs Europe \\
\url{https://jzr99.github.io/syn4d/}
}

\begin{document}
\maketitle

\begingroup
\renewcommand\thefootnote{*}
\footnotetext{Equal contribution.}
\endgroup

\begin{abstract}
Progress in tasks like 3D reconstruction and tracking of dynamic scenes from monocular video is constrained by the scarcity of high-quality datasets with dense, complete, and accurate geometric annotations.
To address this limitation, we introduce \dataname, a multiview synthetic dataset of dynamic scenes that includes ground-truth camera motion, depth maps, dense tracking, and parametric human pose annotations.
A key feature of \dataname is the ability to unproject \emph{any} pixel into 3D to \emph{any} time and to \emph{any} camera.
We conduct extensive evaluations across multiple downstream tasks to demonstrate the utility and effectiveness of the proposed dataset, including 4D scene reconstruction, 3D point tracking, geometry-aware camera retargeting, and human pose estimation.
The experimental results highlight \dataname's potential to facilitate research in dynamic scene understanding and spatiotemporal modeling.
\end{abstract}

\input{figs/teaser.tex}

\input{sec/01_introduction.tex}
\input{sec/02_related.tex}
\input{sec/03_method.tex}
\input{sec/04_experiments.tex}

\input{sec/05_conclusions.tex}

\paragraph*{Acknowledgements}

The authors of this work were supported by Clarendon Scholarship, NTU SUG-NAP, NRF-NRFF17-2025-0009, ERC 101001212-UNION, and EPSRC EP/Z001811/1 SYN3D. %

\bibliographystyle{splncs04}
\bibliography{main,vedaldi_general,vedaldi_specific,chuanxia_general}

\input{sec/supp}

\end{document}

%% file: preamble.tex
\usepackage[accsupp]{axessibility}
\usepackage{array}
\usepackage{booktabs}
\usepackage{eccvabbrv}
\usepackage{xspace}
\usepackage{graphicx}
\usepackage{multirow}
\usepackage[pagebackref,breaklinks,colorlinks]{hyperref}
\usepackage{orcidlink}

\newcommand{\dataname}{\textsc{Syn4D}\xspace}

%% file: figs/teaser.tex
\begin{figure}[h]
\centering
\vspace{-0.6cm}
\includegraphics[width=\textwidth]{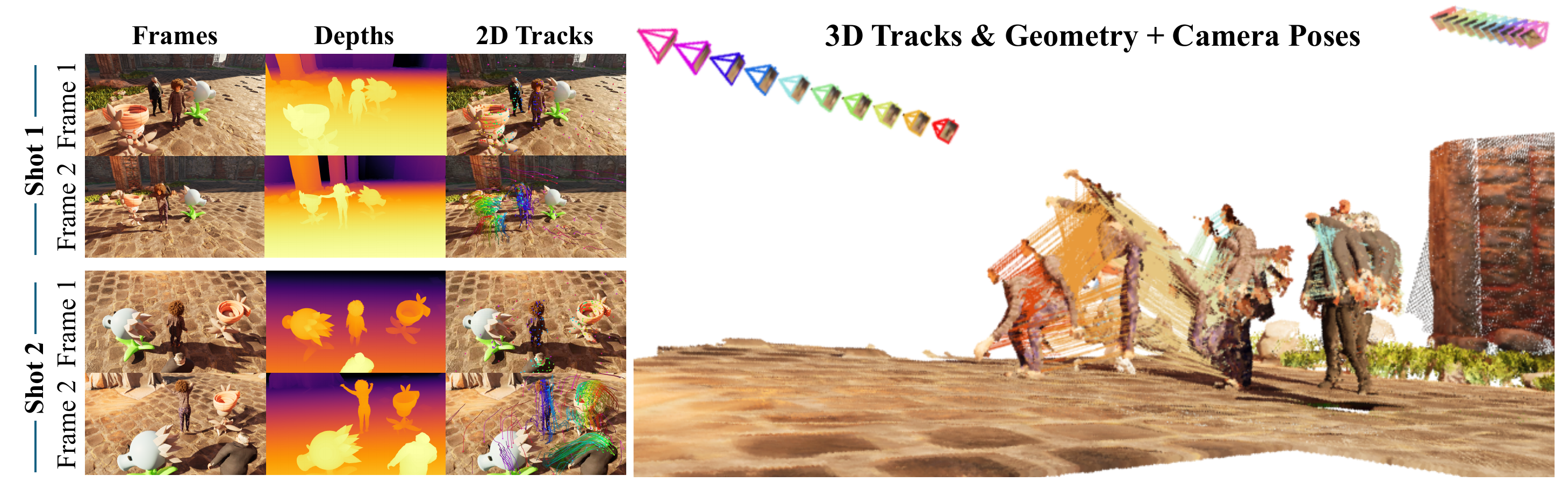}
\caption{\textbf{\dataname} is a large-scale synthetic dataset designed for a set of 4D tasks, including camera pose estimation, depth estimation, dynamic 3D scene reconstruction, 2D/3D tracking, human pose estimation, and novel-view synthesis.
On the left, we visualize two shots with two synchronized frames for each shot.
On the right, we unproject depth maps and visualize the camera trajectories, geometry, as well as 3D tracking results for both shots.}%
\label{fig:teaser}
\vspace{-0.8cm}
\end{figure}

%% file: sec/01_introduction.tex
\section{Introduction}%
\label{sec:intro}

Much of the progress in computer vision is driven by the development of large, general-purpose neural networks, such as transformers, with research focusing on improved learning formulations and data engineering.
3D reconstruction has long resisted this trend, but recent works have shown that feed-forward networks~\cite{wang24dust3r:,wang25vggt,wang26pi3:,keetha26mapanything:,lin25depth,wang26vggt-omega}  can reconstruct 3D scenes,
from one, a few, or hundreds of views,
with performance comparable to traditional optimization-based pipelines like structure-from-motion (SfM).
Furthermore, learning-based approaches can dramatically outperform geometry-based counterparts on more ambiguous tasks, such as reconstructing 3D motion from videos (4D reconstruction)~\cite{feng25st4rtrack:,sucar25dynamic,jin24stereo4d:,sucar2025vdpm,luo20264rc,zhu2026motioncrafter,zhang2026efficiently,han25d2ust3r:,zhang25pomato:}, which have eluded traditional approaches for half a century.
These problems, in fact, \emph{require} the use of statistical priors~\cite{torresani08nonrigid}, which are better captured by learning-based approaches.

So why has machine learning taken so long to become competitive with methods like SfM\@?
We argue that a main reason has been the lack of large-scale datasets with reliable geometric annotations.
Works like the aforementioned~\cite{wang24dust3r:,wang25vggt,wang26pi3:,keetha26mapanything:,lin25depth,wang26vggt-omega} dedicate significant effort to collecting, curating, and annotating synthetic and real data.
However, these datasets and annotations are largely private, and often only contain \emph{static} scenes, which are insufficient for 4D reconstruction.
This is a significant bottleneck to further progress.

In this paper, we address this gap and contribute \textbf{\dataname}, a synthetic dataset designed to advance learning-based 4D reconstruction, tracking, novel-view synthesis, and related tasks (\cref{fig:teaser}).
We mitigate the limitations of existing datasets in scale, quality, annotation coverage, and availability.
Synthetic data has proven highly effective in 2D and 3D geometry tasks such as 3D reconstruction~\cite{yao20blendedmvs:} and generation~\cite{deitke23objaverse:}, human pose estimation~\cite{black23bedlam:}, and 2D/3D tracking~\cite{greff22kubric:,zheng23pointodyssey:,karaev24cotracker}.
Conversely, real datasets such as Stereo4D~\cite{jin24stereo4d:}, while valuable, typically provide only sparse and noisy geometric annotations.

We construct \dataname procedurally using Unreal Engine, leveraging the large catalog of high-quality 3D environments available through the Unreal Fab store.
For dynamic content, we extract $1{,}674$ animated 3D assets from Objaverse(-XL)~\cite{deitke23objaverse:,deitke23objaverse-xl:} and $585$ simulated dynamic humans from Bedlam2~\cite{tesch2025bedlam2}. %
We then place these assets at carefully curated locations within each 3D environment, and design diverse camera trajectories, ranging from simple motions to complex movements, to encourage better generalization of models trained on this data.
This combination of professionally designed 3D environments and diverse animated 3D assets\,---\,humans, animals, humanoid robots, and other characters\,---\,allows producing high-quality videos of dynamic scenes, as shown in \cref{fig:sample}.

Our \dataname also comes with comprehensive annotations, including ground-truth camera motion, depth maps, point maps, \emph{dense} 2D and 3D tracking, and parametric human pose annotations, all in a \emph{multiview} setting.
This makes it one of the few datasets that can support a wide variety of multi-view 3D and 4D tasks, from 3D tracking to and novel view synthesis (NVS).
Some datasets like Kubric~\cite{greff22kubric:}, PointOdyssey~\cite{zheng23pointodyssey:}, and Bedlam2~\cite{tesch2025bedlam2} contain dynamic scenes, but they lack multiview information and dense tracking annotations, which are crucial for learning 4D reconstruction.
In particular, this is the first dataset to provide dense and complete 3D tracking annotations in a multiview setting, enabling the recovery of the 3D position of any point in any image at any time (and thus its 2D projection in any other frame and camera), offering an advantage over datasets like PointOdyssey~\cite{zheng23pointodyssey:}.
Because storing this information explicitly is infeasible, we also develop an efficient representation of these dense tracks based on pixel-aligned barycentric maps paired with the corresponding mesh sequences.
This design enables efficient querying of dense tracks within data loaders for training or evaluation.

Overall, \dataname contains \textbf{4.7K multiview video clips}, totaling \textbf{1.4M frames}, with \textbf{dense geometry annotations}.
This is significantly larger than existing 3D datasets that offer dense geometry and motion annotations (\cref{tab:datamain}).

Empirically, we show that training on \dataname significantly improves the performance of off-the-shelf models in geometry-aware NVS, 4D reconstruction and tracking, and human pose estimation.
Because our \dataname contains \emph{multiview} videos with corresponding ground-truth geometry, we also introduce a new task, benchmark, and evaluation metrics for geometry-aware dynamic NVS, considering both the visual quality and geometric consistency of the generated views.
Another unique property of \dataname is its dense tracking annotations, which support new benchmarks and evaluation metrics for 3D tracking.
Finally, we evaluate the performance of state-of-the-art 4D reconstruction and human pose prediction models trained on \dataname.

To summarize our contributions, \dataname is the first publicly available dataset containing multiview videos of dynamic scenes with dense 3D tracking annotations, along with camera images, depth maps, and human pose labels.
We demonstrate that this data can be used to improve the performance of state-of-the-art 2D/3D tracking and 4D reconstruction models.
We further show that the dataset enables training new models, including a geometry-aware multiview diffusion model that can generate novel-view video sequences together with consistent geometry.

%% file: sec/02_related.tex
\section{Related Work}%
\label{sec:related}

\input{tables/related-v1.tex}

\subsection{3D and 4D datasets}%
\label{sec:rel_dataset}

We review datasets related to \dataname with respect to their synthetic \vs real nature and static \vs dynamic content.
As shown in \cref{tab:datamain}, for \emph{realistic} dynamic scenes with \emph{diverse} categories and \emph{multiview dense} geometry and motion annotations, none of the existing datasets can fully meet these requirements.

\paragraph{Synthetic static 3D datasets.}

Synthetic data offers numerous benefits, including ``perfect'' annotations.
However, creating high-quality synthetic 3D data is complex.
In part, this is due to the difficulty of procuring diverse 3D assets, including scenes, characters, and animations, for building useful synthetic scenes.
ShapeNet~\cite{chang15shapenet}, Objaverse(-XL)~\cite{deitke23objaverse:,deitke23objaverse-xl:}, and others~\cite{wu23omniobject3d:} collect a large number of 3D assets from the web, and have significantly advanced the 3D and 4D generators~\cite{liu23zero-1-to-3:,zhang24clay:,xiang25structured,hunyuan3d25hunyuan3d,wu25amodal3r,jiang26mesh4d}, but they are limited mainly to \emph{objects}. %
For \emph{scene-level} datasets, such as BlendedMVS~\cite{yao20blendedmvs:}, Replica~\cite{straub19the-replica}, Structured3D~\cite{zheng2020structured3d}, Hypersim~\cite{roberts21hypersim:}, SUN RGB-D~\cite{song2015sun}, TartanAir~\cite{wang20tartanair:}, and 3D-FRONT~\cite{fu20213d}, they are either relatively small in scale (\eg, Replica), or lack realistic materials (\eg, Structured3D and 3D-FRONT) and ground-truth geometry (\eg, Hypersim and SUN RGB-D).

Some authors also explore procedural generators to create infinite 3D scenes.
For example, InfiniGen~\cite{raistrick23infinite} provides an engine to build any number of 3D scenes procedurally but contains few animations and highly stylised content.
InfiniGen Indoors~\cite{raistrick24infinigen} constructs indoor environments instead.

\paragraph{Synthetic dynamic 3D datasets.}

Instead of providing only static geometry, more relevant to our work are dynamic synthetic datasets.
To diversify the content, datasets like Sintel~\cite{butler12a-naturalistic} and Spring~\cite{mehl23spring:} utilize free open source Blender movies.
Others, like FlyingThings3D~\cite{mayer16a-large} and Kubric~\cite{greff22kubric:} opt for ``domain randomization'', tossing random objects from ShapeNet~\cite{chang15shapenet} and Google Scanned Object (GSO)~\cite{downs22google}, respectively.
AI2-THOR~\cite{kolve17ai2-thor:} and ProcTHOR~\cite{deitke22procthor:} provide instead manually-authored interactive 3D environments.
BlenderProc2~\cite{denninger23blenderproc2:} can also be used to create data similar to FlyingThings3D via procedural generation.
Other datasets tap video games for content.
A notable example is VIPER~\cite{richter17playing}, which extracts depth, cameras,
and semantic and instance segmentation from GTA\@;
another is JTA~\cite{fabbri18learning}, which further extracts body poses.
Others again, use domain-specific resources;
for instance Virtual KITTI~\cite{gaidon16virtual,cabon20virtual} builds on the CARLA~\cite{dosovitskiy17carla:} simulator, providing synthetic renderings, depth, and semantic segmentation.
SYNTHIA~\cite{ros16the-synthia}, Synscapes~\cite{wrenninge18synscapes:}, SURREAL~\cite{varol17learning}, PreSIL~\cite{hurl19precise} and SEED4D~\cite{kastingschafer25seed4d:} 
also focus on the driving/urban domain.
While they provide some dynamic content, most of them use only a few dynamic objects, or are restricted to specific categories such as cars and pedestrians, or rigid objects.
Recent BEDLAM~\cite{black23bedlam:,tesch2025bedlam2} provides more complex and realistic animations, but focuses only on humans.
DynamicReplica~\cite{karaev23dynamicstereo} introduces other characters, but only $13$ in total.
In contrast, our \dataname combines $1{,}674$ selected animated objects from Objaverse~\cite{deitke23objaverse:,deitke23objaverse-xl:}
and $585$ 3D human assets from BEDLAM~\cite{black23bedlam:,tesch2025bedlam2},
along with $30$ large-scale 3D scene assets we acquired from Sketch Fab,
containing natural-looking clutter, complex illumination, and varied geometry and materials.

For the annotations, some datasets, such as Sintel, Spring, FlyingThings3D, TartanAir, SEED4D, and Kubric, provide all or some of the scene flow and stereo.
Although scene flow is equivalent to dense 3D tracking, it is only available for consecutive frames of a video.
Stereo provides two viewpoints that are close together.
Others like BlendedMVS~\cite{yao20blendedmvs:} and SYNTHIA~\cite{ros16the-synthia} provide multiview geometry, but only for static scenes or a few dynamic objects.
PointOdyssey~\cite{zheng23pointodyssey:} and Dynamic Replica~\cite{karaev23dynamicstereo} provide long-term tracking, but only for very \emph{sparse} points.
In contrast, our \dataname provides \emph{dense multiview} 3D tracking for any pair of frames, including those far apart in time and space.
One direct competitor is Kubric~\cite{greff22kubric:}, which can be used to create dynamic scenes with dense multiview geometry and tracking.
However, it contains only rigid objects that fall to the ground, which lacks realism.
It also lacks 3D background scenes.
Our \dataname, on the other hand, provides a realistic rendering with Unreal Engine 5 and contains a much more diverse and realistic set of 3D assets, scenes, and lighting.

\paragraph{Real static 3D datasets.}

Here, we briefly review the real 3D datasets.
ABO~\cite{collins22abo:}, GSO~\cite{downs22google}, and OmniObject3D~\cite{wu23omniobject3d:} provide object scans.
For the indoor scene datasets,
notable examples like Matterport3D~\cite{Matterport3D}, SceneNN~\cite{ankur15scenenet:} and its follow-up SceneNet RGB-D~\cite{mccormac17scenenet}, ScanNet($++$)~\cite{dai17scannet:,yeshwanth23scannet:} and ARKitScenes~\cite{dehghan2021arkitscenes}, provide RGB-D scans of real indoor scenes, using different RGBD sensors and scanning protocols. %
However, the resulting data is often noisy, and the geometry is incomplete.
Besides, Gibson~\cite{xia18gibson} and Habitat~\cite{habitat19iccv,szot21habitat} build 3D environment based on scanning real 3D scenes.
Although the holes are largely filled, the resulting geometry remains inaccurate.
In parallel, many outdoor datasets, such as KITTI~\cite{geiger13vision}, SemanticKITTI~\cite{behley2019iccv}, nuScenes~\cite{caesar2020nuscenes}, Waymo Open Dataset~\cite{sun20scalability}, KITTI-360~\cite{liao2022kitti}, and Argoverse2 (3D)~\cite{Argoverse2} provide large-scale outdoor scans,
but with very sparse geometry.

\paragraph{Real dynamic 3D datasets.}

For dynamic scenes, real datasets are even more limited.
While TUM RGB-D~\cite{Sturm2012ABF}, ICL-NUIM~\cite{handa:etal:ICRA2014}, and ETH3D~\cite{schops2017multi} provide RGB-D videos, they do not contain real dynamic objects,
but only static scenes with camera motion.
Middlebury~\cite{baker11a-database} builds stereo and optical flow benchmarks using real videos, but they are very small in scale and contain only a few rigid objects.
To record real scenes with dynamic content, Panoptic Studio~\cite{joo15panoptic,joo19panoptic} builds a dome with more than 500 cameras to capture multi-person interactions in a studio.
3DPW~\cite{marcard18recovering} scans 3D body shapes and poses of people in outdoor scenes using IMUs.
PROX~\cite{PROX:2019} captures humans interacting with scanned indoor scenes using a multi-camera setup.
However, these datasets are limited to human interactions and do not contain other dynamic objects or complex scenes, making them less suitable for general 4D reconstruction and tracking research.

\subsection{Applications}

\paragraph{4D reconstruction and tracking.}%
\label{sec:rel_4D}

Recent contributions like DUSt3R~\cite{wang24dust3r:}, and its follow-ups~\cite{wang25continuous,wang25vggt,wang26pi3:,keetha26mapanything:,lin25depth} have made significant progress on \emph{feed-forward} 3D reconstruction by using both real and synthetic large scale datasets for training.
These methods have recently been extended for feed-forward 4D reconstruction and tracking across time~\cite{feng25st4rtrack:,sucar25dynamic,jin24stereo4d:,sucar2025vdpm,karhade25any4d:,luo264rc:}, working by extending the 3D reconstructors to dynamic scenes.
In parallel, some works~\cite{jiang25geo4d,zhang2025GVFD,jiang26mesh4d,ShapeGen4D,zhu26motioncrafter:} have explored the potential of using pre-trained video generators~\cite{blattmann23stable,hu25depthcrafter:} for this task.
Another category of methods has tackled 4D reconstruction as an extension of 2D point tracking by combining it with video depth predictions~\cite{SpatialTracker, zhang25tapip3d:}.

\paragraph{Camera retargeting.}%
\label{sec:rel_novel}

Camera retargeting aims to generate content from arbitrary user-specified viewpoints.
Early works, such as Zero-1-to-3~\cite{liu23zero-1-to-3:} and follow-ups~\cite{zheng24free3d,voleti2024sv3d,chen24mvsplat360,zhou2025stable,szymanowicz25bolt3d:,gao24cat3d:},
present camera-pose-conditioned diffusion models for NVS, but they are limited to \emph{static} scenes.
Camera Dolly~\cite{van2024generative} extends a similar pipeline to \emph{dynamic} scenes by finetuning SVD~\cite{blattmann23stable} with multiview videos rendered by Kubric~\cite{greff22kubric:}.
Recently, there are several works~\cite{bahmani2025vd3d,he2024cameractrl,bai2024syncammaster,bai25recammaster:,yu25trajectorycrafter:,xie26lavr} targeting \emph{long-term} videos and \emph{large-scale} camera motions, but they mainly focus on different architectures and conditions.
In contrast, we provide \emph{multiview} geometry-grounded videos, which can be used not only for novel-view appearance synthesis but also for novel-geometry synthesis.

\paragraph{Human pose estimation.}%
\label{sec:rel_human}

Human pose and shape estimation from images typically fits parametric body models—SMPL~\cite{smpl2015} and SMPL-X~\cite{pavlakos19expressive} either from cropped single-person detections~\cite{kanazawa18end-to-end,kolotouros19learning,goel23humans,patel2025camerahmr,yin2025smplest}, achieving accurate per-person estimates but discarding scene context for occlusion and interaction, or leveraging the full image for richer, scene-aware estimation~\cite{sun2021monocular,jiang2020coherent,sun2022putting,baradel2024multi,wang2025prompthmr,su2025sat,wang2025towards}.
A key finding is that combining diverse synthetic datasets (\eg, BEDLAM~\cite{black23bedlam:}, AGORA~\cite{patel2021agora}) with real-image collections is critical for accurate estimation at scale.
\dataname advances this direction by providing SMPL-X annotations rendered in high-quality, geometrically diverse 3D environments with multiview cameras.

%% file: tables/related-v1.tex
\begin{table}[t]
\newcommand{\Y}{{\color{green}\checkmark}}
\newcommand{\y}{{\color{yellow}$\bullet$}}
\newcommand{\N}{{\color{red}$\times$}}
\newcolumntype{H}{>{\setbox0=\hbox\bgroup}c<{\egroup}@{}}
\centering
\resizebox{\linewidth}{!}{\input{tables/related-contents.tex}}
\caption{\textbf{Comparisons with previous synthetic datasets with geometry annotation.} *We only count the number of frames that have geometry annotation.
All datasets include camera and depth/disparity annotations. Other annotations may include Optical Flow (OF), Scene Flow (SF), Instance Segmentation (IS) consistent across frames, Long-term Point Tracking (Track), Human Pose (HP), noted in the table. A  \y\xspace means that the annotation can be derived in part from the ones provided (for instance SF can be derived from depth, OF and cameras   up to occlusions).}%
\label{tab:datamain}
\end{table}

%% file: tables/related-contents.tex
\begin{tabular}{@{}lcccccccccccH@{}}
\toprule
\textbf{Dataset} & \textbf{Camera} & \textbf{Track} & \textbf{HP} & \textbf{SF} & \textbf{OF} & \textbf{IS} & \textbf{Capt.} & \textbf{Engine} & \textbf{\# Dyn. Obj}. & \textbf{\# Clips} & \textbf{\# Frames} & \textbf{Notes} \\
Sintel~\cite{butler12a-naturalistic} & Stereo & \N & \N & \N & \Y & \Y & \N & Blender & Few others & 35 & 1.6K & Blender Movie \\
Spring~\cite{mehl23spring:} & Stereo & \N & \N & \Y & \Y & \Y & \N & Blender & Few others & 47 & 6K & Blender Movie \\
FlyingThings3D~\cite{mayer16a-large} & Stereo & \N & \N & \Y & \Y & \Y & \N & Custom & Many Rigid & 2.2K & 35K &  \\
Kubric (MOVi-F)~\cite{greff22kubric:} & Mono & Dense & \N & \Y & \Y & \Y & \N & Blender & Many Rigid & 4K & 96K &  \\
Falling Things~\cite{tremblay18falling} & Stereo & \N & \N & \N & \N & \Y & \N & Unreal & 21(Rigid) & 3.5K & 61K &  \\
VIPER~\cite{richter17playing} & Mono & \N & \N & \y & \Y & \Y & \N & GTA V & Cars/Pedestrians & N/A & 254K &  \\
JTA~\cite{fabbri18learning} & Mono & \N & \Y & \y & \y & \Y & \N & GTA V & Cars/Pedestrians & 512 & 461K &  \\
TartanAir~\cite{wang20tartanair:} & Stereo & \N & \N & \y & \Y & \Y & \N & Unreal & Few others & 1037 & 1M &  \\
SceneNet RGB-D~\cite{mccormac17scenenet} & Mono & \N & \N & \y & \Y & \Y & \N & Custom & None & 15K & 5M & SceneNet, ShapeNet \\
Virtual KITTI~\cite{gaidon16virtual,cabon20virtual} & Mono & \N & \N & \y & \Y & \Y & \N & Unity & Cars/Pedestrians & 35 & 17K & CARLA \\
Virtual KITTI 2~\cite{cabon20virtual} & Stereo & \N & \N & \Y & \Y & \Y & \N & Unity & Cars/Pedestrians & 50 & 21K & CARLA \\
BlendedMVS~\cite{yao20blendedmvs:} & Multi & \N & \N & \y & \y & \N & \N & Custom & None & 113 & 18K & 3D scans \\
Dynamic Replica~\cite{karaev2023dynamicstereo} & Stereo & Sparse & \N & \y & \Y & \Y & \N & Blender & 375(humans)+13(others) & 524 & 169K & Replica, 375 humans, 13 other animals \\
PointOdyssey~\cite{zheng23pointodyssey:} & Multi & Sparse & \N & \N & \N & \Y & \N & Blender & 42(humans)+7(others) & 104 & 216K & 42 humans + 7 others \\
BEDLAM2$^\ast$~\cite{black23bedlam:,tesch2025bedlam2} & Mono & \N & \Y & \N & \N & \N & \N & Unreal & 4K(humans) & 12K & 3.5M & N/A (per-frame human pose) \\
SYNTHIA~\cite{ros16the-synthia} & Multi & \N & \N & \N & \N & \N & \N & Unity & Cars/Pedestrians & 4 & 200K & Fixed multi-camera \\
SEED4D~\cite{kastingschafer25seed4d:} & Multi & \N & \N & \y & \Y & \Y & \N & Unreal & Cars/Pedestrians & 10.5K & 16.8M & Dense / 4D dynamic supervision (designed for dynamic 4D) \\
SURREAL~\cite{varol17learning} & Mono & \N & \Y & \y & \y & \Y & \N & Blender & 930(humans) & 67K & 6.5M & Dense (per-frame pose, depth, seg) \\
Synscapes~\cite{wrenninge18synscapes:} & Mono & \N & \N & \N & \N & \Y & \N & Custom & None & Still images & 25K & Custom procedural generator \\
\midrule
Kubric (Ours) & Multi & Dense & \N & \Y & \Y & \Y & Global & Blender & Many Rigid & 5.6K & 274K & As Kubric \\
\dataname  (Ours) & Multi & Dense & \Y & \Y & \Y & \Y & G.+Local & Unreal & 585(humans)+1674(others) & 4.7K & 1.4M & See paper \\
\bottomrule
\end{tabular}

%% file: sec/03_method.tex
\section{Method}%
\label{sec:method}

We first formulate the dataset content in \cref{sec:content}.
Then, in \cref{sec:dpm}, we describe our key contribution to efficiently store the dense tracking annotations.
In \cref{sec:filtering}, we describe how we construct the dataset, including asset filtering and camera motion design.
We generate global and local captions using a vision-language model, as discussed in \cref{sec:captioning}.
Finally, we propose a geometry-aware multiview diffusion model trained on our \dataname dataset in \cref{sec:diffusion}.

\input{figs/sample}

\subsection{Dataset content}%
\label{sec:content}

\dataname consists of a collection $\mathcal{S}$ of clips.
Each clip is obtained by randomly combining a 3D environment with various dynamic 3D objects, then rendering and annotating them, as further detailed in \cref{sec:construction}.
The clip contains information captured by $C$ \emph{cameras} over $T$ frames.
First, the cameras produce RGB frames $I_i \in \mathbb{R}^{3\times H\times W}$.
Each frame thus has a \emph{camera index} $c_i\in \{ 0,\dots,C-1 \}$ and a \emph{time index} $t_i \in \{0,\dots,T-1\}$.
In addition to the RGB frame, each camera also comes with its \emph{parameters} (extrinsics and intrinsics) $\pi_i$.
Because there is one frame per camera and per time, the index $i$ ranges from 1 to $N = TC$.

A unique feature of our dataset is the ability to \emph{track any point}.
This means that, given a pixel $u\in \{ 0,\dots,H-1 \} \times \{ 0,\dots,W-1 \}$ in an image $I_i$, we can tell the 3D position of the corresponding point at any time, as well as its 2D projection in any other camera.
Formally, this information is captured by the \emph{dynamic point maps} (DPMs)~\cite{sucar25dynamic,sucar26v-dpm}
$$
P_i(\pi_k,t_j) \in \mathbb{R}^{3\times H\times W}.
$$
There is one DPM $P_i$ for each image $I_i$.
If $u\in \{ 0,\dots,H-1 \} \times \{ 0,\dots,W-1 \}$ is a pixel, then $P_i(\pi_k,t_j)(u) \in \mathbb{R}^3$ is the 3D location that the physical point corresponding to pixel $I_i(u)$ has at time $t_j$ (which may differ from the time $t_i$ of the image),
expressed in reference frame $\pi_k$ (which can also differ from the reference frame $\pi_i$ of the image's camera).
Finally, we also provide instance segmentation maps $S_i$ that associate each pixel $u$ of an image with a unique ID, distinguishing different dynamic objects in the clip.

To summarize, for each clip our dataset provides tuples $\{(I_i,P_i,S_i, t_i,\pi_i)\}_{i=1}^N$ describing the appearance and dense 3D geometry and motion of the scene.
In addition, depth maps $D_i$ are derived by taking the $z$ channel of the corresponding point maps $P_i(\pi_i,t_i)$, and they are also included in the DPMs.
Additional metadata includes captions for each clip, both global and local (one every 81 frames), describing the clip's content, useful for tasks like learning video generators.

\subsection{An efficient dynamic point map representation}%
\label{sec:dpm}

To the best of our knowledge, our dataset is the first to provide \emph{multi-view} dense tracking annotations for general dynamic objects.
In part, this can be explained by the difficulty of storing such annotations.
For example, PointOdyssey~\cite{zheng23pointodyssey:} explicitly stores sparse tracking annotations, and it would be theoretically possible, but impractical, to store one such track for each pixel in every image.

With our notation, dense tracks are captured by DPMs $P_i(\pi_j,t_k)$.
Each DPM is a $H\times W$ image with three components per pixel and is indexed by $i$, $\pi_j$, and $t_k$.
The index $i$ enumerates the $N$ images in the clip, whereas $\pi_j$ and $t_k$ index viewpoints and times.
There are $N$ possible viewpoints $\pi_j$, one for each image (camera-time combination) in the clip, and $T$ possible times.
Hence, all DPMs together form a $(3\times H\times W) \times N \times N \times T$ tensor, which has space complexity $\mathcal{O}(HWT^3C^2)$ (by definition $N=TC$).
A saving comes from the fact that point maps that differ only by the reference frame $\pi$ are related by a known rigid transformation; \ie, given $Q_i(t_k) = P_i(\pi_0,t_k)$ for a fixed reference frame $\pi_0$ (\eg, that of camera $c_0$ at time $t_0$), we can find $P_i(\pi_j,t_k)$ from $Q_i(t_k)$ by applying the known rigid transformation between $\pi_0$ and $\pi_j$.
This reduces the space complexity to $\mathcal{O}(HWT^2C)$, which is much better, but still impractical.
For instance, if $H=W=512$, $T=300$, and $C=8$, and assuming four bytes per scalar, we would have to store 2.1TiB of information for just one clip!
This should be compared to the 7GiB required to store the (raw) RGB frames, which is negligible in comparison.

Intuitively, this representation is highly redundant because each 3D point trajectory is represented multiple times, once for each image that observes it.
Internally, Kubric~\cite{greff22kubric:} takes advantage of the rigidity of the objects it contains to represent tracks efficiently, but this is not possible here due to the non-rigid deformations of the objects.

We solve this problem as follows.
First, we note that all 3D surfaces in the clip can be represented as an animated triangular mesh, the union of the meshes of all clip objects.
Let this mesh be represented by a collection of time-varying vertices $q_v(t)$, $v=1,\dots,V$ and (fixed) triangular faces $F \subset \{ 1,\dots,V \}^3$ that are triplets of such vertices.
Assume that the vertices are expressed in the reference frame $\pi_0$.
Given a pixel $u$ in image $I_i$, we can recover the 3D point $Q_i(t)(u)$ by finding the triangular face $f=(f_1,f_2,f_3) \in F$ that contains that pixel and computing the corresponding barycentric coordinates $\alpha=(\alpha_1,\alpha_2,\alpha_3)\in\Delta_3$.
Here $\Delta_3\subset[0,1]^3$ is the 3-simplex, \ie, the set of triples of non-negative numbers that sum to one.
Then, for all $t=0,\dots,T-1$, we have:
\begin{equation}\label{eq:barycentric}
Q_i(t)(u) = \alpha_1 q_{f_1}(t) + \alpha_2 q_{f_2}(t) + \alpha_3 q_{f_3}(t).
\end{equation}
Hence, for each pixel $u$, we only need to store four scalars (one index to identify the face $f$ in the collection $F$, and three barycentric coordinates $\alpha$).
Together with the $3 \times V \times T$ scalars for the vertex trajectories, this has complexity $\mathcal{O}(HWTC + VT)$, which is much more manageable.
In the example above, with $V=100K$ vertices, we would only need 9.3GiB to store all DPMs.
In practice, further significant reductions are possible by packing the data in fewer bits and noting that the information is required only for pixels belonging to dynamic objects, which constitute a small fraction of the total.

Note also that the decoding is highly efficient: to recover the track that passes through pixel $u$ in frame $I_i$, \ie, the sequence $(Q_i(t))_{t=0}^{T-1}$, we linearly combine vectors $(q_f(t))_{t=0}^{T-1}$ as shown in \cref{eq:barycentric}.

\paragraph{Extraction.}

Unreal Engine has no rendering pass that can directly produce the quantities $f$ and $\alpha$ discussed above, so we compute them in pre-processing.
Given the depth map $D\in \mathbb{R}^{H\times W}$ (which Unreal provides) and camera rotation $R\in \mathbb{R}^{3\times 3}$, position $o\in \mathbb{R}^{3}$, and intrinsics $K \in \mathbb{R}^{3\times 3}$ for an image $I$, we recover the 3D coordinate of the point at pixel $u$ as
$
\bar q(u) = R^{-1}K^{-1}D(u) \, (u,1)^\top + o.
$
Then, the face $f(u)$ and coordinates $\alpha(u)$ for that pixel are given by:
$$
(f(u),\alpha(u)) = \operatornamewithlimits{argmin}_{f \in F, \alpha \in \Delta_3}
\| \bar q(u) - \alpha_1 q_{f_1}(t) - \alpha_2 q_{f_2}(t) - \alpha_3 q_{f_3}(t)\|.
$$

\subsection{Dataset construction}%
\label{sec:construction}

Next, we describe how we generate our synthetic clips, including their 3D content, cameras, and captions.

\paragraph{Content.}%
\label{sec:filtering}

We construct scenes by combining 3D environments and dynamic assets and randomly placing them according to manually defined rules.

For the \emph{environments}, we manually select (and purchase) 30 3D scenes from the Fab store and choose an appropriate location in each scene as the root position for placing our dynamic objects and humans.

For the \emph{objects}, unlike Bedlam2, which contains only humans, we include a more diverse set of dynamic objects, such as robots, animals, and monsters, from the Objaverse dataset.
However, the 4D assets from Objaverse contain noisy, low-quality animations.
We therefore first render each animated object from a bird's-eye view to filter out objects with large deformations or fast motion by analyzing the segmentation masks.
Specifically, we compare the intersection-over-union (IoU) between adjacent frames to estimate motion speed and filter out frames with low IoU and extreme deformations.

For the \emph{composition}, we randomly choose 1--3 dynamic objects from the filtered Objaverse dataset and one human from the Bedlam2 released assets.
As in Bedlam2, we use the ground-occupancy map to randomly initialize the layout to avoid collisions.

For \emph{illumination}, we use the Lumen real-time global illumination system introduced in UE5 for nearby movable light sources, and precomputed baked lightmap textures for static lights.

\paragraph{Camera motion.}%
\label{sec:camera}

\begin{figure*}[t]
\centering
\includegraphics[width=\textwidth]{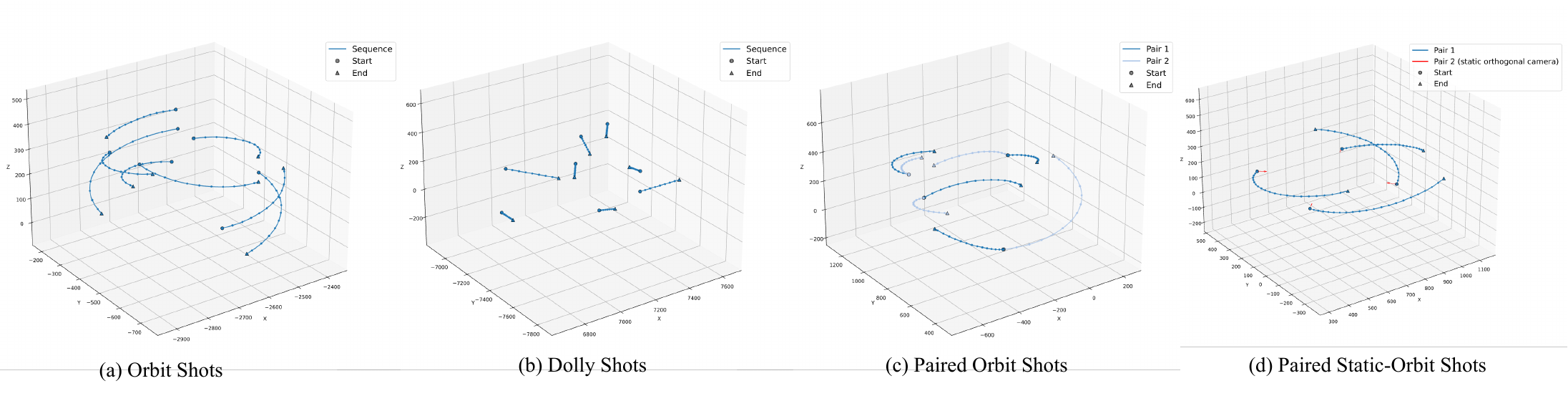}
\caption{\textbf{Sample camera motion in our dataset.}
}%
\label{fig:syn4d_camera}
\end{figure*}

Another key design decision is the camera optics, placement, and motion.
For the camera \emph{intrinsics}, we cover a broad range of horizontal fields of view from 39.6$^{\circ}$ to 90$^{\circ}$.
We dynamically adjust the focal length range depending on whether the scene is indoor or outdoor to better mimic real-world shots.
For the camera \emph{extrinsics}, we generate a variety of camera-motion combinations: static, tracking, dolly, and orbit.
Similar to Bedlam2, we also apply synthetic Perlin-noise camera shake to the camera extrinsics on top of these motions.
The majority of camera motions are orbits, allowing us to capture the full dynamic scene with fewer shots.
We use keyframes to define the extrinsic changes in Unreal Sequencer.
For the orbit shots, we randomly select the initial camera position in spherical coordinates relative to the camera root within a specified range.
We then randomly generate the delta values for the radial distance, polar angle, and azimuth angle between the first and last keyframes.

To obtain a \emph{multiview} dataset, we generate eight different cameras per clip.
As shown in \cref{fig:syn4d_camera}(a,b), for some scenes we generate eight independently sampled orbit and dolly shots to maximize the coverage of the dynamic scene. %
For the others, we generate paired shots.
Specifically, in \cref{fig:syn4d_camera}(c), we generate two paired orbit shots that start from a shared frame. %
As shown in \cref{fig:syn4d_camera}(d), we first generate four orthogonal static shots and then use them as starting points for another four orbit shots. %
These different multiview camera patterns facilitate training the multiview diffusion model to generalize to various source and target camera configurations.

\subsection{Captioning}%
\label{sec:captioning}

We use Tarsier2--7B~\cite{yuan2025tarsier2advancinglargevisionlanguage} to caption our \dataname dataset.
Since each video sequence contains around 300 frames, we provide captions at two levels of granularity.
We sample 16 frames from the entire sequence to generate one global caption.
We then split the video into clips of 81 frames and sample 32 frames from each clip to generate one local caption.
These local and global captions provide users with flexibility when sampling videos of varying lengths for training.

\subsection{Geometry-aware multiview diffusion model}%
\label{sec:diffusion}

In the experiments, we test how our new dataset can help in a variety of tasks, from 4D reconstruction to new-view synthesis.
We also explore a new task uniquely supported by our data, namely the simultaneous synthesis of a novel view and the corresponding 4D geometry (\ie, the 3D points and their motion).

Given a source video
$I \in \mathbb{R}^{3 \times H \times W \times T}$
and its corresponding point maps
$(P_i(\pi_0,t_i))_{i=0}^{T-1} \in \mathbb{R}^{3 \times H \times W \times T}$,
we synthesize a target video $I' \in \mathbb{R}^{3 \times H \times W \times T}$ and its corresponding point maps
$(P'_i(\pi_0,t_i))_{i=0}^{T-1} \in \mathbb{R}^{3 \times H \times W \times T}$
from a new set of cameras $\pi'=(\pi'_i)_{i=0}^{T-1}$.
We start from ReCamMaster~\cite{bai25recammaster:}, which takes as input a video $I$ and the target camera trajectory $\pi'$ to generate a new video $I'$ seen under $\pi'$.
We then extend it to take $P$ as input and to output $P'$.
Because ReCamMaster uses latent diffusion, we adapt their video encoder
$
\mathcal{L}_\text{vid}=\mathcal{E}(V) \in \mathbb{R}^{H\times W\times T\times d}
$
into a DPM encoder
$
\mathcal{L}_\text{DPM}=\mathcal{E}(P) \in \mathbb{R}^{H\times W\times T\times d}.
$
We then reuse their model architecture but spatially concatenate the video latents as
$
\mathcal{Z}
=
[\mathcal{L}_\text{vid} ; \mathcal{L}_{\text{DPM}}]
\in
\mathbb{R}^{H\times (2W) \times T \times d}.
$
With spatial concatenation, we can leverage the pretrained diffusion prior for the geometry-aware novel-view synthesis task without introducing new model parameters.

%% file: figs/sample.tex
\begin{figure}[tb!]
\centering
\includegraphics[width=\textwidth]{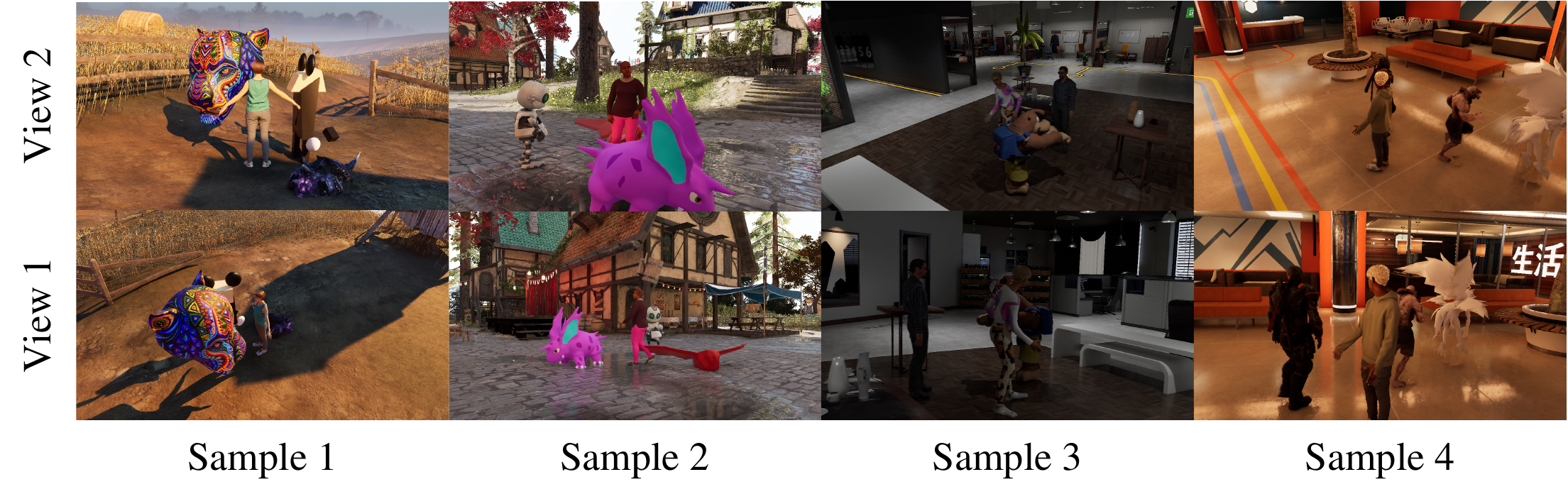}
\caption{\textbf{\dataname} is a multiview synthetic dataset with diverse dynamic objects, including humans, animals, humanoid robots, and other characters.}%
\label{fig:sample}
\end{figure}

%% file: sec/04_experiments.tex
\section{Experiments}%
\label{sec:experiments}

To evaluate the quality of \dataname and its effectiveness for improving state-of-the-art models on a set of 4D attributes, we conduct thorough experiments on various tasks.
We first report results on our proposed task of geometry-aware novel view synthesis in \cref{sec:nvs}.
Then, in \cref{sec:4d}, we train a state-of-the-art 4D reconstruction and tracking model, showing that our dataset supports a broad set of tasks, including 3D tracking, video depth estimation, camera pose estimation, and multi-view reconstruction.
Finally, in \cref{sec:pose}, we fine-tune a state-of-the-art human pose estimator on our dataset, further demonstrating its utility for human pose estimation.

\subsection{Geometry-aware multiview diffusion model}%
\label{sec:nvs}

\paragraph{Metrics.}

Following baseline ReCamMaster~\cite{bai25recammaster:},
we report the FVD~\cite{unterthiner2019accurategenerativemodelsvideo} and CLIP-V~\cite{radford21learning} to measure the temporal consistency of generated novel views and their alignment with the source video, respectively.
Except for the visual appearance,
we further estimate camera trajectories from generated views using Back on Track~\cite{chen25back},
align them to the ground-truth conditioned trajectories with the Umeyama algorithm, and report three standard metrics: Absolute Translation Error (ATE), Relative Translation Error (RPE-T), and Relative Rotation Error (RPE-R).
For geometry evaluation, direct pixel-aligned absolute relative error is unsuitable because the generated novel views are not pixel-aligned with ground-truth target videos.
We therefore convert point maps to RGB and evaluate them with CLIP-V and FVD, denoted as CLIP-V-P and FVD-P, respectively.

\paragraph{Experiment setup.}

We fine-tune our geometry-aware multiview generator on the Kubric and \dataname datasets.
We evaluate the model on our in-house benchmark, which includes 280 video pairs with scenes and dynamic objects that are completely unseen during training.

\input{tables/nvs_geo}

\paragraph{Results.}

As shown in \cref{tab:nvs_geo}, our model trained on \dataname outperforms the model trained on Kubric in all visual and geometric metric.
It achieves camera accuracy comparable to that of the Kubric baseline, suggesting that future versions of the dataset would benefit from more diverse paired camera patterns.

\subsection{4D reconstruction and tracking}%
\label{sec:4d}

Here, we further evaluate whether training with \dataname improves learning-based 4D reconstruction.
Quantitative results (\cref{tab:4drecon_tracking,tab:4drecon_camrecon,tab:4drecon_depth}) show that co-training the 4D reconstructor with \dataname consistently surpasses its original performance across multiple 4D tasks, including 3D tracking, camera pose estimation, multi-view 3D reconstruction,
and video depth estimation.
This shows that \dataname provides valuable geometric supervision for learning spatiotemporal scene representations.
Details are discussed in the following.

\paragraph{Metrics.}

We employ task-specific metrics for each dimension of 4D reconstruction.
For \emph{3D tracking}, we use the Average Percentage of Points (APD) within a threshold and the End-Point Error (EPE).
For \emph{camera pose estimation}, we report Absolute Translation Error (ATE) and Relative Pose Error (RPE) for both translation and rotation.
Multi-view \emph{point maps reconstruction} is assessed using Accuracy (Acc), Completeness (Comp), and Normal Consistency (NC).
\emph{Video depth estimation} is evaluated using the Absolute Relative Error (Rel) and threshold accuracy ($\delta < 1.25$), under both scale and scale-and-shift alignment.

\paragraph{Experiment setup.}

We adopt 4RC~\cite{luo264rc:} as our baseline.
It jointly predicts camera poses, dense geometry, and motion from monocular videos, in a single feed-forward pass.
We use the \emph{de facto} architecture and exactly the same training protocol as 4RC,
while augmenting the training data with the \dataname dataset.

For evaluation, we follow the standard protocols used in prior work~\cite{luo264rc:, wang24dust3r:, wang25continuous, stream3r2025}.
Sparse 3D point tracking is evaluated on Aerial Digital Twin (ADT)~\cite{pan2023ariadigitaltwinnew} and Panoptic Studio (PStudio)~\cite{pstudio} from TAPVid-3D~\cite{koppula24tapvid-3d:}, which provide sparse trajectory annotations.
To further assess the performance on dense tracking, we additionally evaluate on two rendered test sets from \dataname, Soviet and Warehouse, each containing 50 sequences.
For each sequence, we sample a 24-frame clip and evaluate dense trajectories starting from the first frame.
We also evaluate camera pose estimation on Sintel~\cite{butler12a-naturalistic}, TUM-dynamics~\cite{Sturm2012ABF}, and ScanNet~\cite{dai17scannet:}, video depth estimation on Sintel, Bonn~\cite{Bonn}, and KITTI~\cite{geiger13vision}, and multi-view 3D reconstruction on 7-Scenes~\cite{Shotton_2013_CVPR} and NRGBD~\cite{Azinovic_2022_CVPR}.

\input{tables/4drecon_tracking.tex}
\input{tables/4drecon_camrecon.tex}

\paragraph{Results.}

We report quantitative results in~\cref{tab:4drecon_tracking,tab:4drecon_camrecon,tab:4drecon_depth}.
Training with \dataname consistently improves performance across all evaluated tasks,
\emph{without requiring any specific architecture modification.}
Notably, clear gains are observed in 3D tracking, showing that improvements from training data are significant,
which suggests another important direction in 3D research, \ie, building large-scale datasets.
A similar conclusion is also observed on camera estimation (comparable or better), multi-view 3D reconstruction, and video depth estimation,
where co-training with our large-scale \dataname consistently improves performance.
These consistent improvements demonstrate the effectiveness of our \dataname and motivate investing in large-scale 4D data creation in future work.

\paragraph{Qualitative results.}

\begin{figure}[tb!]
\centering
\includegraphics[width=\linewidth]{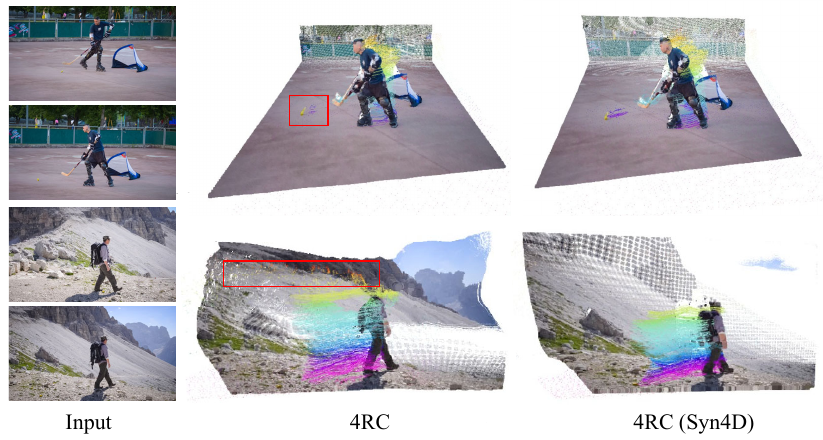}
\caption{\textbf{Qualitative comparison of in-the-wild 4RC~\cite{luo264rc:} tracking trained w/o and w/ \dataname.} Training with \dataname yields more accurate object tracks (\eg, the ball) and more stable distant geometry (\eg, the sky and far mountains), owing to the more complete and diverse annotations in \dataname.
}%
\label{fig:syn4d_track_compare}
\end{figure}

We further provide a qualitative comparison of in-the-wild 4D reconstruction and tracking in \cref{fig:syn4d_track_compare}.
Training the 4D reconstructor with \dataname yields more accurate object tracks (\eg, the ball) and more stable distant geometry (\eg, the sky and far mountains),
owing to the more complete and diverse annotations in \dataname.
This indicates that \dataname improves not only the quantitative metrics but also the perceptual quality of reconstruction and tracking on real-world videos.

\input{tables/4drecon_depth.tex}

\subsection{Human pose estimation}%
\label{sec:pose}

Each scene in \dataname also contains a single human with ground-truth SMPL-X~\cite{pavlakos19expressive} body pose and shape annotations.
We thus convert these to SMPL~\cite{smpl2015} via the optimization-based refitting provided by SMPL-X,
enabling feed-forward training with SMPL-based methods.

\paragraph{Metrics.}

We report three standard human mesh recovery metrics:
Per-Joint Position Error (MPJPE),
Procrustes-Aligned MPJPE (PA-MPJPE),
and Per-Vertex Error (PVE)~\cite{black23bedlam:},
all in millimeters ($\downarrow$).
We evaluate on the test sets of 3DPW~\cite{von2018recovering}, Hi4D~\cite{yin2023hi4d},
and the training set of CHI3D~\cite{fieraru2020three},
three challenging real-image benchmarks.
Note that,
none of the models is trained or fine-tuned on the benchmark datasets.

\paragraph{Experiment setup.}

We adopt MA-HMR~\cite{wang2025towards} as our baseline, a state-of-the-art single-stage multi-person SMPL recovery method.
It is trained with AGORA~\cite{patel2021agora}, BEDLAM~\cite{black23bedlam:}, and DTO-Humans~\cite{wang2025towards}.
We fine-tune MA-HMR's final checkpoint on the original three datasets plus \dataname (MA-HMR+\dataname) for six epochs.
To ensure a fair comparison, we also fine-tune on the original three datasets alone (MA-HMR+Cont) for exactly the same number of epochs.
We filter out frames in which the human is heavily occluded using the segmentation masks.
We use a learning rate of  $10^{-5}$ and a total batch size of 128.

\paragraph{Results.}

As shown in \cref{tab:human_pose}, MA-HMR+Cont yields similar performance to the baseline, confirming that additional training epochs alone does not account for the gain.
Adding \dataname (MA-HMR+\dataname) consistently improves all three metrics across all three benchmarks, demonstrating that our synthetic human annotations provide a useful complement to existing training corpora.
We attribute the improvement to the diverse occlusion patterns in \dataname, which are well-suited to MA-HMR's one-stage design.

\input{tables/human_pose}

%% file: tables/nvs_geo.tex
\begin{table}[tb!]
\centering
\resizebox{\linewidth}{!}{%
\begin{tabular}{@{}l cc ccc cc@{}}
\toprule
\multirow{2}*{Method} & \multicolumn{2}{c}{\textbf{Visual Quality}} & \multicolumn{3}{c}{\textbf{Camera Accuracy}} & \multicolumn{2}{c}{\textbf{Geometry Quality}} \\ 
\cmidrule(lr){2-3}
\cmidrule(lr){4-6}
\cmidrule(lr){7-8}
& CLIP-V~$\uparrow$ & FVD~$\downarrow$ & ATE~$\downarrow$ & RPE trans~$\downarrow$ & RPE rot~$\downarrow$ & CLIP-V-P~$\uparrow$ & FVD-P~$\downarrow$  \\ 
\midrule
Ours (Kubric) & 0.643 & 631 & \textbf{0.064} & 0.023 & 0.328 & 0.757 & 229 \\

Ours (\dataname) & \textbf{0.740} & \textbf{452} & 0.070 & \textbf{0.021} & \textbf{0.272} & \textbf{0.816} & \textbf{139}  \\

\bottomrule
\end{tabular}
}
\caption{\textbf{Quantitative evaluation} for geometry-aware novel view synthesis on \dataname evaluation benchmark.}%
\label{tab:nvs_geo}
\end{table}

%% file: tables/4drecon_tracking.tex
\begin{table*}[tb!]
\centering
\renewcommand{\arraystretch}{1.0}

\setlength{\tabcolsep}{3pt} 
\begin{tabular}{@{}l cc cc cc cc@{}}
\toprule
\multirow{3}{*}{\textbf{Method}} & \multicolumn{4}{c}{\textbf{Sparse Point Tracking}} & \multicolumn{4}{c}{\textbf{Dense Tracking}} \\
\cmidrule(lr){2-5} \cmidrule(lr){6-9}
 & \multicolumn{2}{c}{ADT~\cite{pan2023ariadigitaltwinnew}} & \multicolumn{2}{c}{PStudio~\cite{pstudio}} & \multicolumn{2}{c}{Soviet} & \multicolumn{2}{c}{Warehouse} \\
\cmidrule(lr){2-3} \cmidrule(lr){4-5} \cmidrule(lr){6-7} \cmidrule(lr){8-9}
 & APD $\uparrow$ & EPE $\downarrow$
 & APD $\uparrow$ & EPE $\downarrow$
 & APD $\uparrow$ & EPE $\downarrow$
 & APD $\uparrow$ & EPE $\downarrow$ \\
\midrule

4RC
& 87.82 & 0.1480 & 87.32 & 0.1304 & 58.35 & 3.8458 & 79.07 & 0.3302 \\

4RC (\dataname)
& \textbf{89.44} & \textbf{0.1258} & \textbf{88.10} & \textbf{0.1276} & \textbf{74.46} & \textbf{1.8934} & \textbf{88.79} & \textbf{0.1915} \\

\bottomrule
\end{tabular}
\caption{\textbf{Quantitative evaluation for 3D tracking.}
We compare the baseline 4RC~\cite{luo20264rc} with the model retrained with the original dataset plus \dataname.
Notably, co-training with \dataname yields significant improvement on both \emph{sparse} and \emph{dense} 3D tracking, with the identical experimental setting.
}%
\label{tab:4drecon_tracking}
\end{table*}

%% file: tables/4drecon_camrecon.tex
\begin{table*}[tb!]
\centering
\renewcommand{\arraystretch}{1.0}
\renewcommand{\tabcolsep}{1.1mm}
\resizebox{\textwidth}{!}{%
\begin{tabular}{@{}l ccc ccc ccc ccc ccc@{}}
\toprule
\multirow{3}{*}{\textbf{Method}} & \multicolumn{9}{c}{\textbf{Camera Pose Estimation}} & \multicolumn{6}{c}{\textbf{Multi-View 3D Reconstruction}} \\
\cmidrule(lr){2-10} \cmidrule(lr){11-16}
 & \multicolumn{3}{c}{Sintel~\cite{butler12a-naturalistic}} & \multicolumn{3}{c}{TUM-dynamics~\cite{Sturm2012ABF}} & \multicolumn{3}{c}{ScanNet~\cite{dai17scannet:}} & \multicolumn{3}{c}{7-Scenes~\cite{Shotton_2013_CVPR}} & \multicolumn{3}{c}{NRGBD~\cite{Azinovic_2022_CVPR}} \\
\cmidrule(lr){2-4} \cmidrule(lr){5-7} \cmidrule(lr){8-10} \cmidrule(lr){11-13} \cmidrule(lr){14-16}
 & ATE $\downarrow$ & RPE$_\text{t}$ $\downarrow$ & RPE$_\text{r}$ $\downarrow$
 & ATE $\downarrow$ & RPE$_\text{t}$ $\downarrow$ & RPE$_\text{r}$ $\downarrow$
 & ATE $\downarrow$ & RPE$_\text{t}$ $\downarrow$ & RPE$_\text{r}$ $\downarrow$
 & Acc $\downarrow$ & Comp $\downarrow$ & NC $\uparrow$
 & Acc $\downarrow$ & Comp $\downarrow$ & NC $\uparrow$ \\
\midrule

4RC 
& 0.144 & 0.053 & 0.430
& \textbf{0.010} & \textbf{0.008} & \textbf{0.314}
& 0.032 & \textbf{0.012} & 0.437
& 0.034 & 0.051 & 0.783
& 0.036 & 0.034 & 0.912 \\

4RC (\dataname)
& \textbf{0.076} & \textbf{0.040} & \textbf{0.302}
& 0.012 & 0.010 & 0.325
& \textbf{0.032} & 0.013 & \textbf{0.384}
& \textbf{0.031} & \textbf{0.043} & \textbf{0.791}
& \textbf{0.029} & \textbf{0.032} & \textbf{0.924} \\

\bottomrule
\end{tabular}
}
\caption{\textbf{Quantitative evaluation for camera pose estimation and multi-view 3D reconstruction.}
Co-training with \dataname significantly improves the 3D reconstruction performances on all instantiations,
while achieves comparable or better performances on camera pose estimation.
}%
\label{tab:4drecon_camrecon}
\end{table*}

%% file: tables/4drecon_depth.tex
\begin{table}[tb!]
\centering
\renewcommand{\arraystretch}{1.0}
\resizebox{\columnwidth}{!}{
\setlength{\tabcolsep}{3pt} 
\begin{tabular}{@{}l @{\hspace{4mm}} c @{\hspace{4mm}} cc cc cc@{}}
\toprule
\multirow{2}{*}{\textbf{Method}} & \multirow{2}{*}{\textbf{Align}} & \multicolumn{2}{c}{\textbf{Sintel}~\cite{butler12a-naturalistic}} & \multicolumn{2}{c}{\textbf{Bonn}~\cite{Bonn}} & \multicolumn{2}{c}{\textbf{KITTI}~\cite{geiger13vision}} \\
\cmidrule(lr){3-4} \cmidrule(lr){5-6} \cmidrule(lr){7-8}
 &  & Rel $\downarrow$ & $\delta < 1.25 \uparrow$ 
 & Rel $\downarrow$ & $\delta < 1.25 \uparrow$ 
 & Rel $\downarrow$ & $\delta < 1.25 \uparrow$ \\
\midrule

4RC
& \multirow{2}{*}{scale}
& 0.311 & 62.2 
& 0.051 & \textbf{97.4} 
& 0.076 & 95.2 \\

4RC (\dataname)
& 
& \textbf{0.211} & \textbf{74.6} 
& \textbf{0.048} & 97.3 
& \textbf{0.071} & \textbf{95.7} \\

\midrule

4RC
& \multirow{2}{*}{scale \& shift}
& 0.249 & 67.0 
& 0.048 & 97.3 
& 0.058 & 95.5 \\

4RC (\dataname)
& 
& \textbf{0.176} & \textbf{76.6} 
& \textbf{0.046} & \textbf{97.3} 
& \textbf{0.057} & \textbf{95.7} \\

\bottomrule
\end{tabular}
}
\caption{\textbf{Quantitative evaluation for video depth estimation.}
We report metrics under both scale-only and scale-and-shift alignments.
Co-training with \dataname improves the performance on almost all datasets across all metrics.
}%
\label{tab:4drecon_depth}
\end{table}

%% file: tables/human_pose.tex
\begin{table}[tb!]
\centering
\renewcommand{\arraystretch}{1.15}
\renewcommand{\tabcolsep}{1.8mm}

\resizebox{\textwidth}{!}{%
\begin{tabular}{@{}l ccc ccc ccc@{}}
\toprule
\multirow{2}*{\textbf{Method}} & \multicolumn{3}{c}{\textbf{Hi4D}~\cite{yin2023hi4d}} & \multicolumn{3}{c}{\textbf{CHI3D}~\cite{fieraru2020three}} & \multicolumn{3}{c}{\textbf{3DPW}~\cite{von2018recovering}} \\ 
\cmidrule(lr){2-4}
\cmidrule(lr){5-7}
\cmidrule(lr){8-10}
& MPJPE~$\downarrow$ & PA-MPJPE~$\downarrow$ & PVE~$\downarrow$ & MPJPE~$\downarrow$ & PA-MPJPE~$\downarrow$ & PVE~$\downarrow$& MPJPE~$\downarrow$ & PA-MPJPE~$\downarrow$ & PVE~$\downarrow$ \\ 

\midrule
MA-HMR  & 58.8 & 43.9 & 73.6 & 47.2 & 31.4 & 55.7 & 63.2 & 40.2 & 73.8\\
MA-HMR+Cont  & 58.7 & 44.1 & 73.2 & 46.9 & 31.4 & 55.5 & 63.0 & 40.0 & 73.4\\
MA-HMR+\dataname & \textbf{57.7} & \textbf{43.0} & \textbf{72.2}& \textbf{46.1} & \textbf{31.1} & \textbf{54.9} & \textbf{62.5} & \textbf{39.6} & \textbf{72.9} \\

\bottomrule

\end{tabular}
}
\caption{\textbf{Quantitative evaluation for human pose estimation.} 
$+$Cont denotes continuing training with the original datasets. $+$\dataname denotes continuing training with the original datasets plus \dataname.}%
\label{tab:human_pose}
\end{table}

%% file: sec/05_conclusions.tex
\section{Conclusions}%
\label{sec:conclusions}

We have introduced \dataname, a large multiview synthetic 4D dataset of dynamic scenes with accurate and complete geometric annotations, including camera motion, depth, dense tracking, and human pose.
Experiments show that \dataname improves performance across many tasks, including 3D tracking, video depth estimation, camera pose estimation, multi-view reconstruction, and human pose estimation.
It also supports new tasks, such as geometry-aware novel view synthesis, which we introduce and explore.
These results demonstrate that \dataname provides effective supervision for learning spatiotemporal scene representations and offers a strong foundation for future research in 4D vision.

%% file: sec/supp.tex
\clearpage

\begin{center}
    {\Large\textbf{Syn4D\@: A Multiview Synthetic 4D Dataset}}

    \vspace{0.1in}
    {{\textit{Supplementary Document}}}

\end{center}

In this \textbf{supplementary document}, we provide additional materials for our main submission.
In the \textbf{supplementary video}, we show more visual results of our dataset.
The \textbf{dataset} will be made publicly available for research purposes.

\section{Dataset statistics}

\begin{figure*}[h]
\centering
\includegraphics[width=\textwidth]{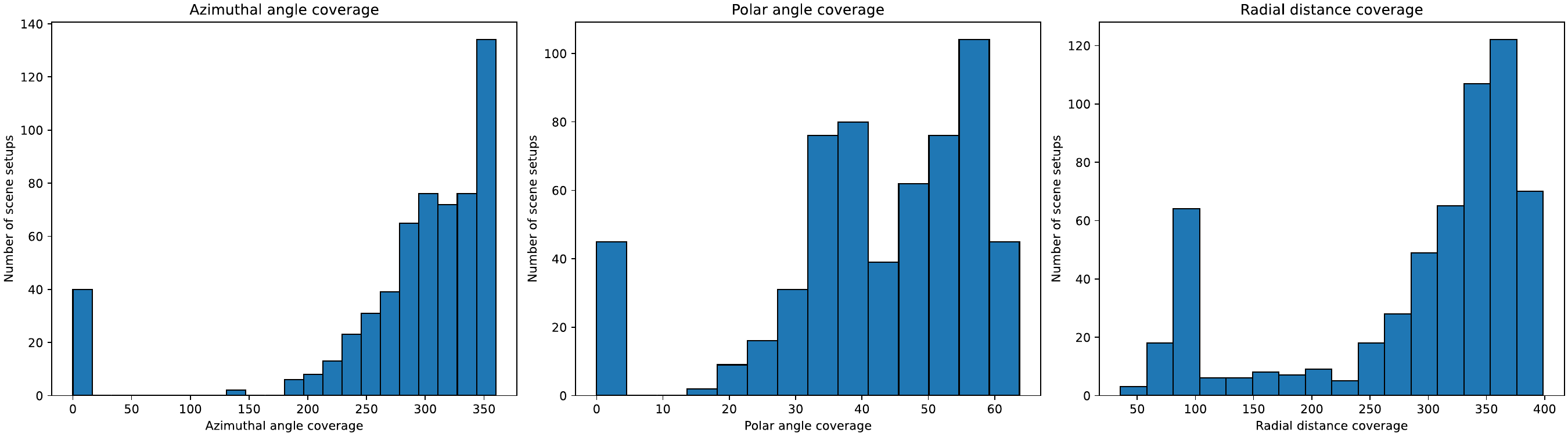}
\caption{\textbf{Dataset camera motion statistics.} For each scene setup, our camera trajectories cover a broad range of azimuthal angles, polar angles, and radial distances.
}%
\label{fig:syn4d_cam_histo}
\end{figure*}

As mentioned in the main paper, for each scene setup we render eight shots to capture the full dynamics of the scene.
As illustrated in \cref{fig:syn4d_cam_histo}, we analyze the coverage of azimuthal angle, polar angle, and radial distance for each scene configuration to demonstrate that the randomly generated orbital camera motion effectively spans the dynamic scene.
Specifically, the majority of scene setups exhibit an azimuthal angle coverage greater than 250$^{\circ}$, a polar angle coverage exceeding 30$^{\circ}$, and a radial distance variation larger than 2.5 m.
These statistics indicate that the rendered camera trajectories provide comprehensive spatial coverage for each scene setup.

\begin{figure*}[h]
\centering
\includegraphics[width=\textwidth]{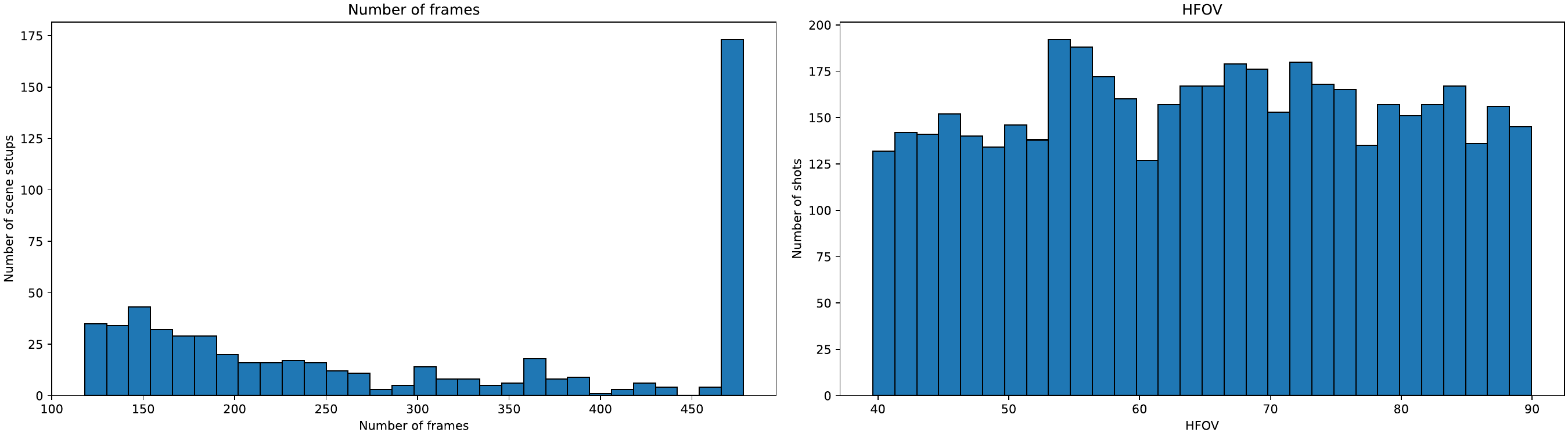}
\caption{\textbf{Dataset frame length and HFOV statistics.} Most shots are longer than 450 frames at 30 FPS and cover a broad range of HFOV\@.
}%
\label{fig:syn4d_hfov_histo}
\end{figure*}

As shown in \cref{fig:syn4d_hfov_histo}, we further analyze the distribution of the number of frames per scene setup and the horizontal field of view (HFOV) for each shot.
Most scenes contain approximately 500 frames captured at 30 FPS\@.
The HFOV for each shot is randomly sampled from a range of 39.6$^{\circ}$ to 90$^{\circ}$, ensuring diverse camera perspectives across the dataset.

\section{Implementation Details}

\subsection{Geometry-aware multiview diffusion}
Our geometry-aware multi-view diffusion model is fine-tuned on top of ReCamMaster~\cite{bai2025recammaster}.
Training is conducted at a resolution of $144 \times 256$ per modality, with a learning rate of $2 \times 10^{-5}$, an effective batch size of $16$, and $49$ frames per sequence.
To enable multi-modality generation, we concatenate different modalities along the width dimension, following~\cite{chen20254dnex}, while keeping the novel-view tokens concatenated along the temporal (frame) dimension, giving $288 \times 512 \times 98$ input shape.
The first $288 \times 512 \times 49$ indicates width-wise concatenated RGB and Point Map, which stay clean during the training, and the second half is diffused according to flow matching diffusion training schedule~\cite{liu23flow}.
Two modality-specific tokens are learned to indicate the modality type.
All models are trained for $15{,}000$ iterations.

\begin{figure*}[h]
\centering
\includegraphics[width=\textwidth]{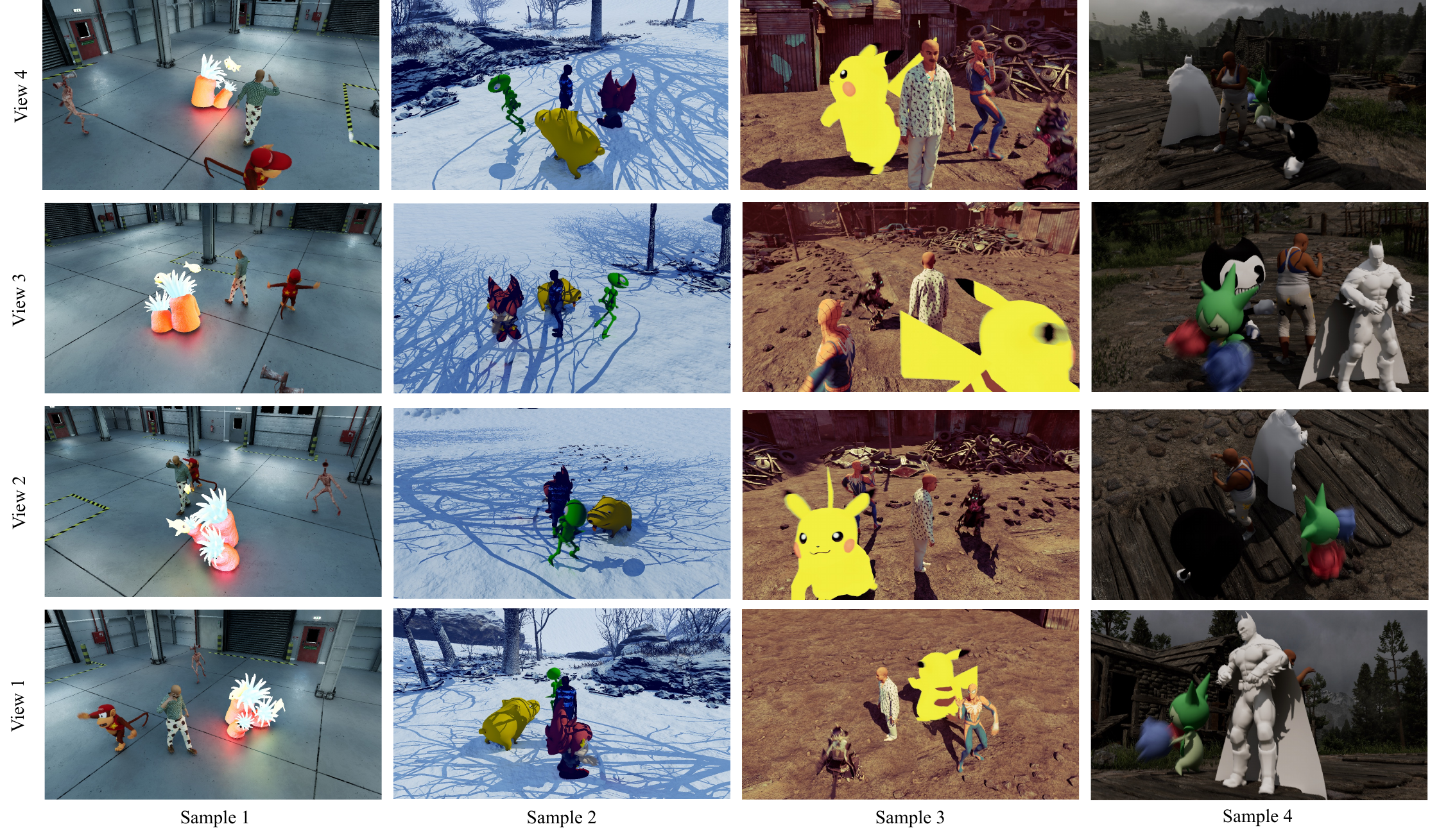}
\caption{\textbf{More samples from our \dataname dataset.} \dataname is a multiview synthetic dataset with diverse dynamic objects, including humans, animals, humanoid robots, and other characters.
}%
\label{fig:syn4d_supp_sample}
\end{figure*}

\subsection{4D reconstruction and tracking}

\newcommand{\yhluo}[1]{\textcolor{blue}{[Yihang: #1]}}

For training 4RC~\cite{luo20264rc}, we use \dataname together with a subset of its original training datasets, including PointOdyssey~\cite{zheng23pointodyssey:}, Dynamic Replica~\cite{karaev2023dynamicstereo}, Waymo~\cite{sun20scalability}, Kubric~\cite{greff22kubric:}, DL3DV~\cite{ling23dl3dv-10k:}, ScanNet++~\cite{yeshwanth23scannet:}, and MVS-Synth~\cite{huang18deepmvs:}. We preserve the original dataset distribution and incorporate \dataname, which constitutes approximately one-sixth of the final training data. For \dataname, we clip the maximum depth value at 300 to avoid distant sky regions, and exclude human hair from supervision as it may introduce floating points. Following 4RC, during training the input images are resized to a randomly sampled resolution, with the longer side up to 504 pixels. We uniformly sample the aspect ratio from $[0.5, 2.0]$, randomly select the sequence length from $2$ to $18$ views, and sample video frames with a random temporal interval between $1$ and $5$ frames. The model is trained for around four days for 50 epochs on 16 A100 GPUs with a batch size of 1 per GPU\@.

\subsection{Human pose estimation}

\paragraph{Occlusion filtering.}

We derive a per-frame visibility mask from Unreal Engine's rendered body and clothing segmentation passes, which are combined into a single binary visibility mask indicating which pixels of the person are unoccluded.
A frame is discarded if any of the following conditions hold:
(i) any segmentation mask file is missing, indicating that the character is fully occluded;
(ii) the bounding-box area computed from the visibility mask is below $10{,}000$ pixels, indicating that the person appears too small or is too distant; or
(iii) the ratio of visible pixels to bounding-box area falls below $0.3$, meaning that less than 30\% of the bounding box is covered by unoccluded body pixels.
After filtering, the \dataname{} subset used for human pose estimation training retains 917K frames.

\paragraph{Training details.}

We fine-tune MA-HMR for 6 epochs using a learning rate of $10^{-5}$, a total batch size of 128, distributed across 4 NVIDIA H100 GPUs.
We use the AdamW optimizer with a weight decay of $10^{-4}$, a cosine learning-rate schedule, and no warmup steps.
Training wall-clock time is approximately 24 hours for MAHMR+Cont and approximately 29 hours for MAHMR+\dataname{}.

\paragraph{Data mixing.}

Training follows the data mixture from MA-HMR~\cite{wang2025towards}.
The three original datasets consist of BEDLAM (6 FPS version, 286K frames), AGORA (14.4K frames), and DTO-Humans (557K frames).
For \dataname{}, we use 23 scenes for the human pose estimation task and apply a 5$\times$ subsampling to the filtered 917K frames, yielding 183K frames --- comparable in scale to the 6 FPS BEDLAM subset.
All datasets are concatenated and shuffled uniformly at each epoch.

\paragraph{Evaluation protocol.}

We evaluate on three benchmarks without any training or fine-tuning on their respective datasets.
For 3DPW~\cite{von2018recovering}, we use the standard test split.
For Hi4D~\cite{yin2023hi4d}, we designate pairs 23, 27, 28, 32, and 37 as the test split.
For CHI3D~\cite{fieraru2020three}, we use the training split downsampled by a factor of 16.
3DPW and Hi4D each comprise approximately 25K test frames; CHI3D contains approximately 16K frames after downsampling.
No model has access to any training data from the evaluation benchmarks.

\section{Additional qualitative results}

\subsection{Dataset}

As shown in \cref{fig:syn4d_supp_sample}, we show more samples from our \dataname dataset.
Our \dataname contains diverse indoor and outdoor scenes together with a large variety of dynamic objects.

\subsection{Geometry-aware multiview diffusion}

\begin{figure*}[h]
\centering
\includegraphics[width=\textwidth]{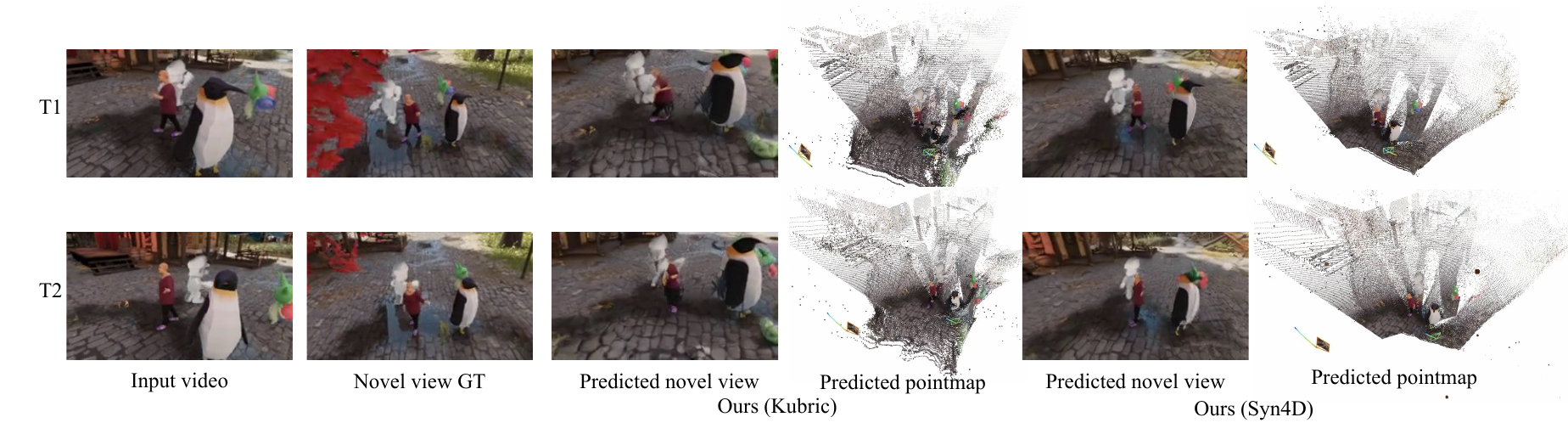}
\caption{\textbf{Qualitative comparison of our geometry-aware multiview diffusion model trained with Kubric and \dataname.} Training with \dataname produces more accurate camera control and aligned geometry across different views.}%
\label{fig:syn4d_supp_geo_compare}
\end{figure*}
As shown in \cref{fig:syn4d_supp_geo_compare}, we provide a qualitative comparison between our geometry-aware multiview diffusion models trained on Kubric and \dataname.
The model trained on \dataname exhibits more accurate camera control and substantially better cross-view geometric consistency than the model trained on Kubric.
These results highlight the importance of our dataset for the newly introduced geometry-aware novel view synthesis task.

\subsection{4D reconstruction and tracking}

\begin{figure*}[h]
\centering
\includegraphics[width=\textwidth]{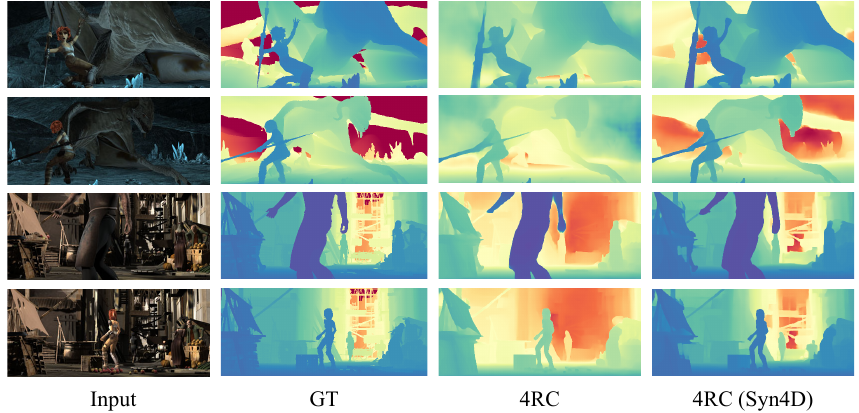}
\caption{\textbf{Qualitative comparison of 4RC~\cite{luo20264rc} trained w/o and w/ \dataname for video depth estimation} on the Sintel~\cite{butler12a-naturalistic} dataset. Training with \dataname produces sharper depth predictions and demonstrates improved geometric understanding.}%
\label{fig:syn4d_supp_depthcomparison}
\end{figure*}

\begin{figure*}[h]
\centering
\includegraphics[width=\textwidth]{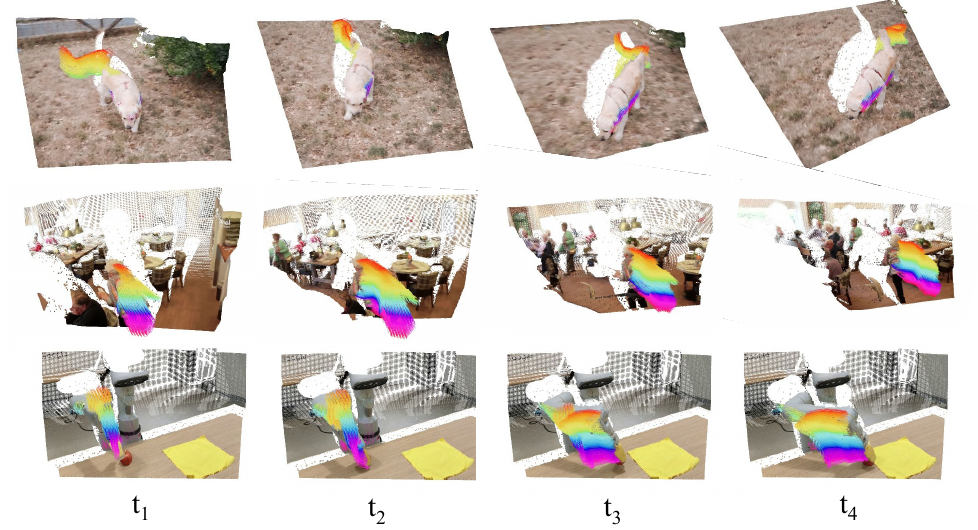}
\caption{\textbf{Visualization of in-the-wild dynamic reconstruction and tracking.} We show the dynamic reconstruction of sampled timestamps from a video, along with dynamic object trajectories rendered as rainbow-colored paths. The model trained with \dataname demonstrates strong performance across diverse in-the-wild scenarios.
}%
\label{fig:syn4d_supp_3dtrack}
\end{figure*}

\Cref{fig:syn4d_supp_depthcomparison} provides a qualitative comparison of depth estimation on the Sintel~\cite{butler12a-naturalistic} benchmark and highlights the improvements gained by integrating \dataname into the 4RC~\cite{luo20264rc} training pipeline.
The model trained with \dataname yields sharper depth predictions and better geometric understanding.
This suggests that \dataname provides the structural diversity necessary for learning high-quality geometry.
Additionally, \cref{fig:syn4d_supp_3dtrack} provides visualization results of the model trained with \dataname on dynamic 3D reconstruction with tracking, which demonstrate its strong performance across diverse in-the-wild scenarios.

\subsection{Human pose estimation}
\input{figs/supp-human_pose}

\Cref{fig:hpe_qualitative} shows two Hi4D examples featuring close-proximity persons under occlusion, comparing MA-HMR and MA-HMR+\dataname{} against the ground truth.
Adding \dataname{} training data produces more accurate body pose estimates in these occluded cases.

\section{Additional quantitative results}

\subsection{Geometry-aware multiview diffusion}

In order to precisely evaluate geometry accuracy for synthesized novel views, we first establish correspondences between source and target images at the same timestep $i$ by using the ground-truth point map $P$ from the source view and $P^{\prime}$ from the target view:
\begin{equation}
\begin{gathered}
\mathcal{M}_{i}=\left \{(u, v) \mid v=\mathrm{NN}(u,P_{i}^{\prime},P_{i}) \text { and } u=\mathrm{NN}(v,P_{i},P_{i}^{\prime})\right \} \\
\text { with } \mathrm{NN}(u,P_{i}^{\prime},P_{i})=\underset{v \in \{0, \ldots, W H\}}{\arg \min }\left \|P_{i,u}^{\prime}-P_{i,v}\right \| .
\end{gathered}
\end{equation}
Here, we retain mutual correspondences $\mathcal{M}_{i}$ for each image pair at frame $i$. Then we only calculate the distance with ground truth for those pixels with valid correspondence between source and target frames to evaluate reconstructed geometry on the visible region instead of on the hallucinated one. The multiview correspondence error (COR) is calculated as follows:
\begin{equation}
L_{cor}=\sum_{i=0}^{T-1} \frac{1}{T|\mathcal{M}_{i}|} \underset{(u,v) \in \mathcal{M}_{i}}{\sum }\left \|P_{i,u}^{\prime}-\hat{P^{\prime}}_{i,u}\right \|,
\end{equation}
where $\hat{P^{\prime}}_{i,u}$ and $P^{\prime}_{i,u}$ are predicted target view point cloud and the ground truth one separately. As shown in \cref{tab:nvs_geo_cor}, the method trained on our \dataname dataset consistently achieves better performance on this metric, highlighting the effectiveness of our dataset. Moreover, the model conditioned on the ground-truth point map outperforms the one conditioned on DA3, further confirming the effectiveness of the point map condition.

\input{tables/nvs_supp}
\input{tables/nvs_geo_master}

As shown in \cref{tab:nvs_geo_master}, we follow the ReCamMaster protocol and evaluate on their test set, which is totally unseen during training.
However, since the ReCamMaster's dataset provides only novel-view videos,
we thus use Depth-Anything-3~\cite{lin25depth} to label pseudo ground truth geometry for evaluation. The model trained on \dataname outperforms the baseline on the zero-shot evaluation benchmark in terms of both visual and geometric quality, further demonstrating the realism and generalizability of the \dataname dataset.

\paragraph{Train/test split and in-domain concern.}

\dataname ships with a train/test split whose backgrounds and objects are \emph{disjoint} from the training set, so the evaluation never reuses training scenes.
To further address fairness with respect to dataset scale, we additionally retrain our model with matched clip counts ($2048$) and frame number (first $49$ frames) for both Kubric and \dataname.
\input{tables/nvs_geo_master_rebuttal}
As shown in \cref{tab:nvs_geo_master_rebuttal}, we evaluate the same Kubric- and \dataname-trained models on the ReCamMaster benchmark (zero-shot for both), and the \dataname-trained model still wins under this matched-budget setting.
We exclude SynCamMaster from this evaluation because it contains only static cameras, whereas ReCamMaster includes more complex camera setups.

%% file: figs/supp-human_pose.tex
\begin{figure*}[ht!]
\centering
\setlength{\tabcolsep}{1pt}
\renewcommand{\arraystretch}{0.5}
\begin{tabular}{cccc}
\includegraphics[width=0.23\textwidth]{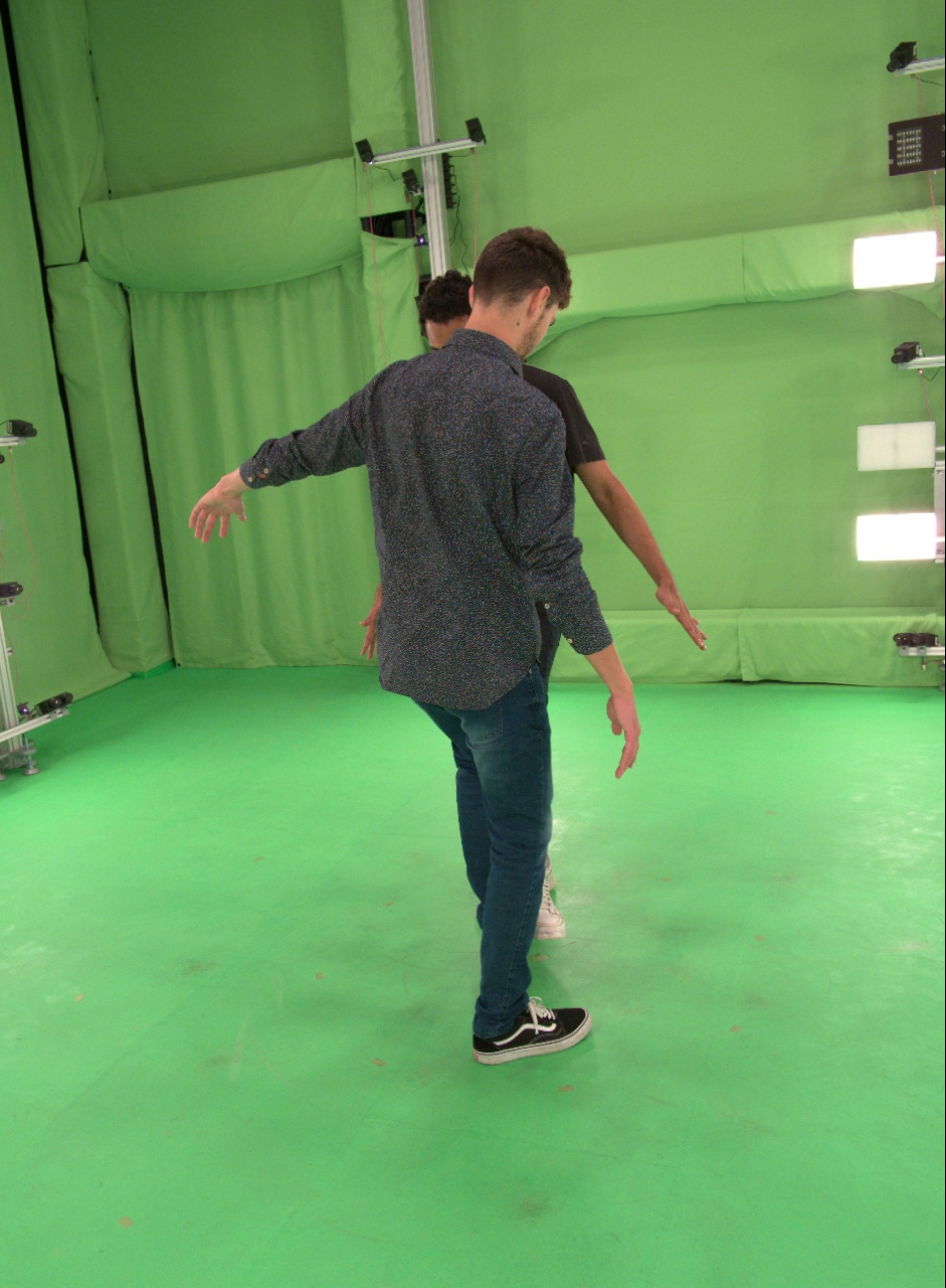} &
\includegraphics[width=0.23\textwidth]{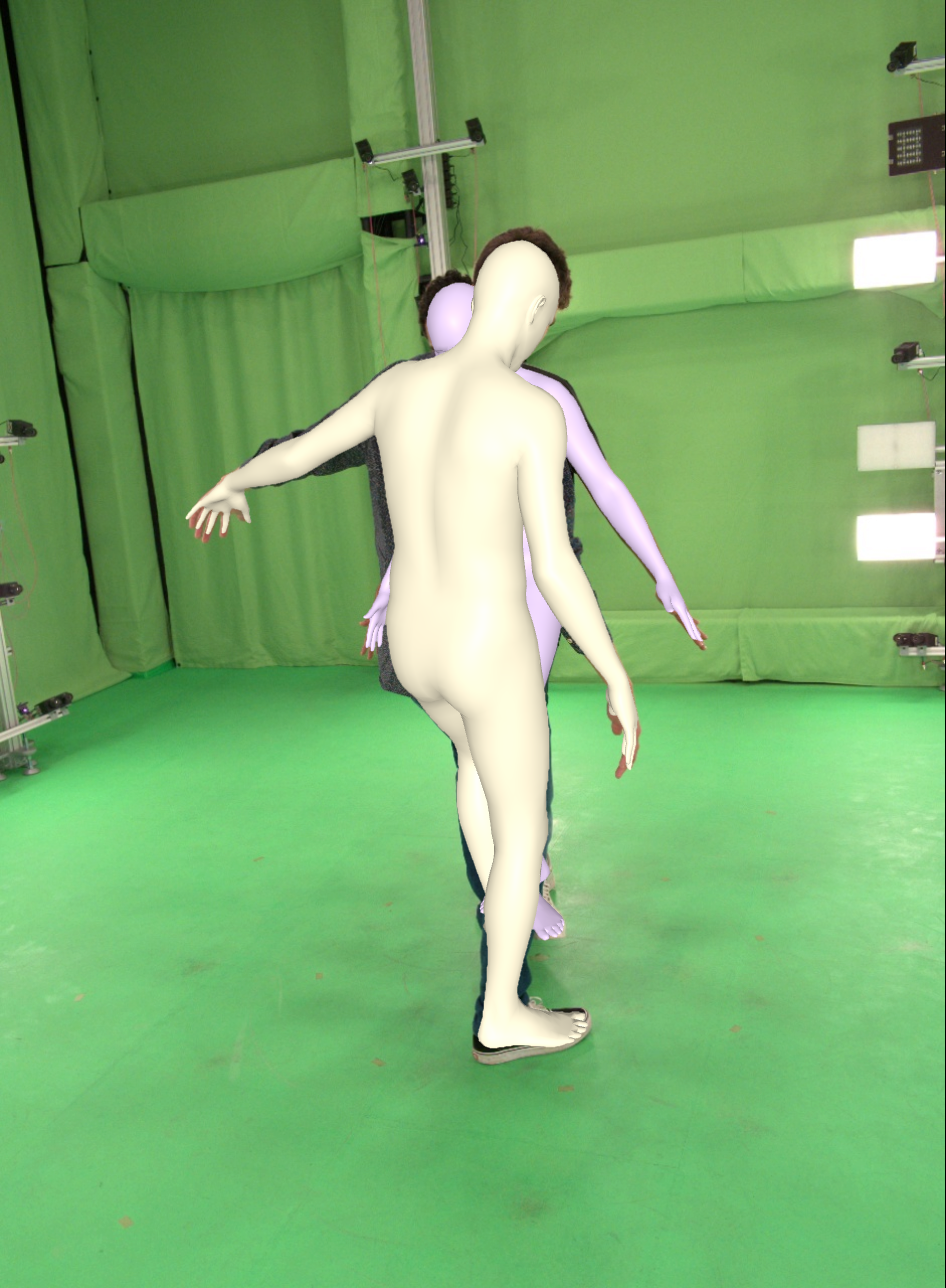} &
\includegraphics[width=0.23\textwidth]{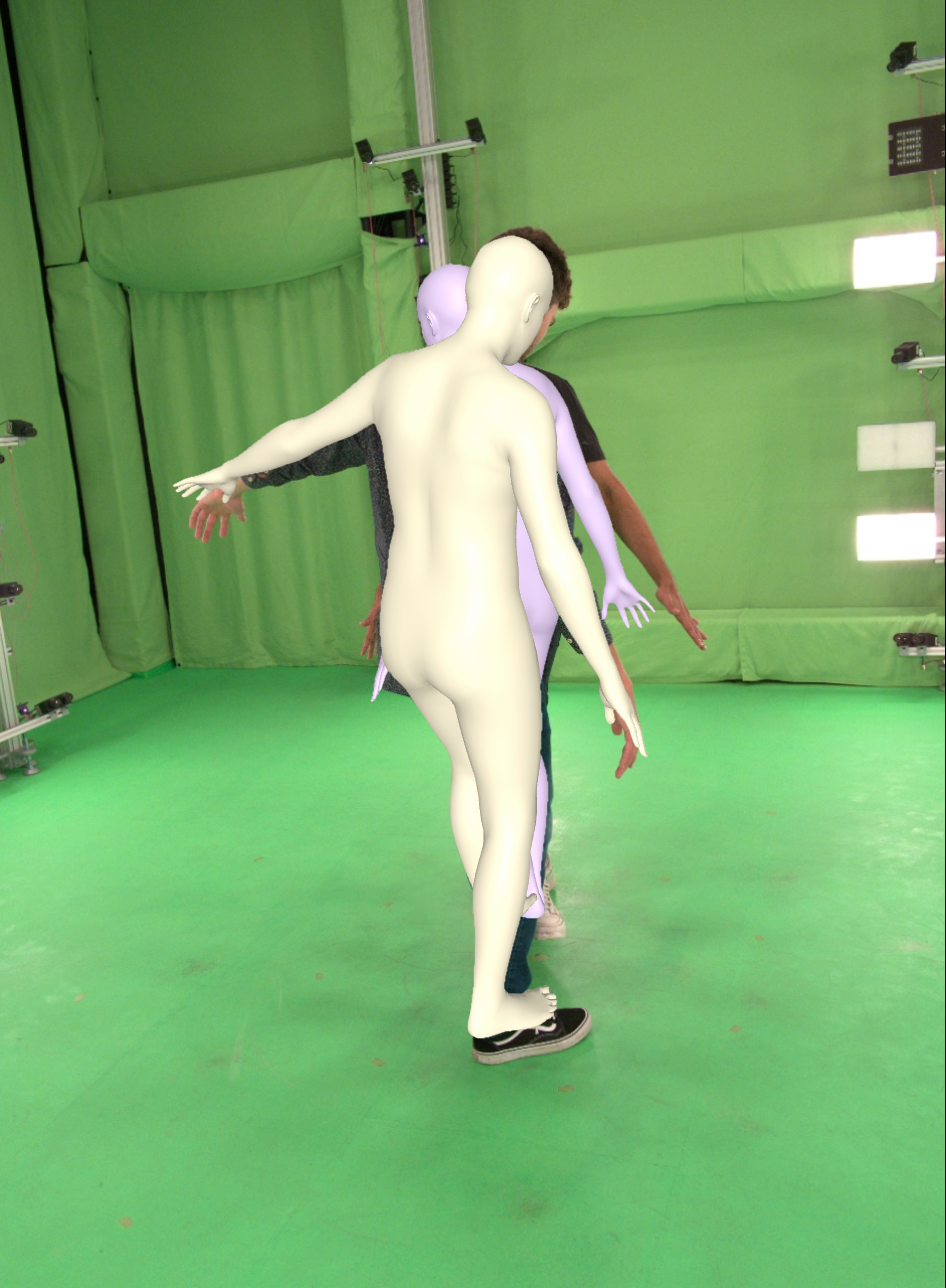} &
\includegraphics[width=0.23\textwidth]{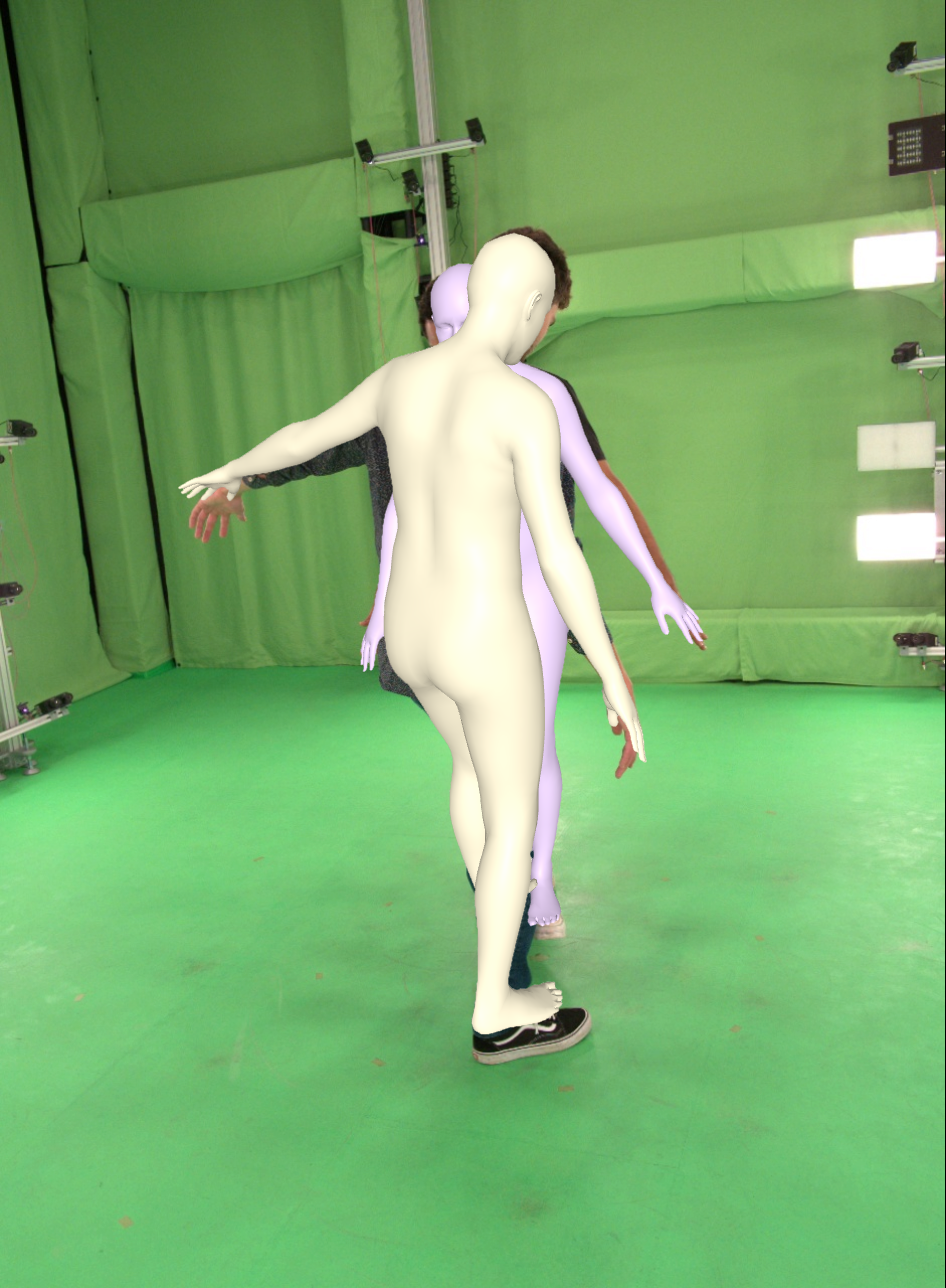} \\[2pt]
\includegraphics[width=0.23\textwidth]{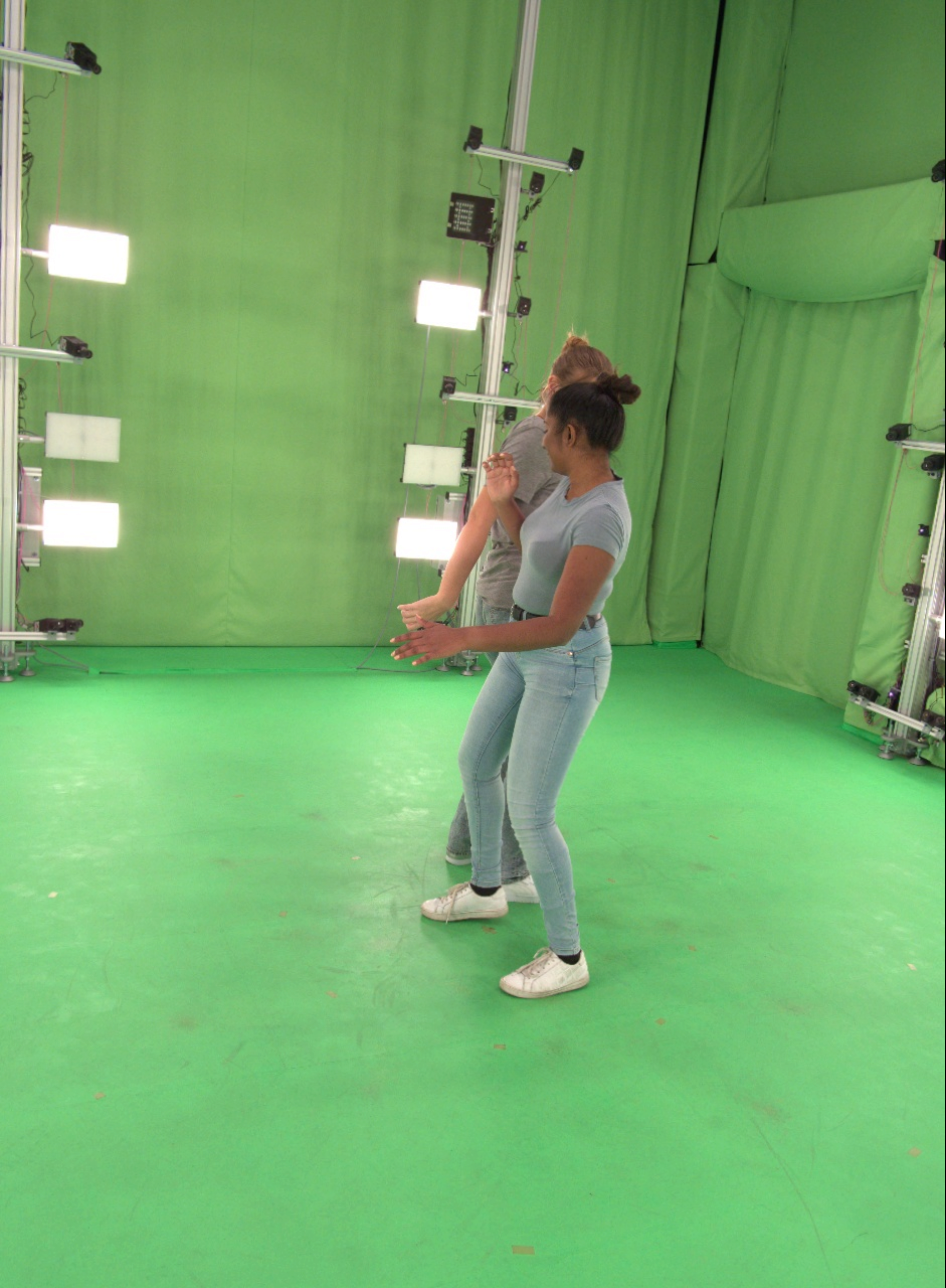} &
\includegraphics[width=0.23\textwidth]{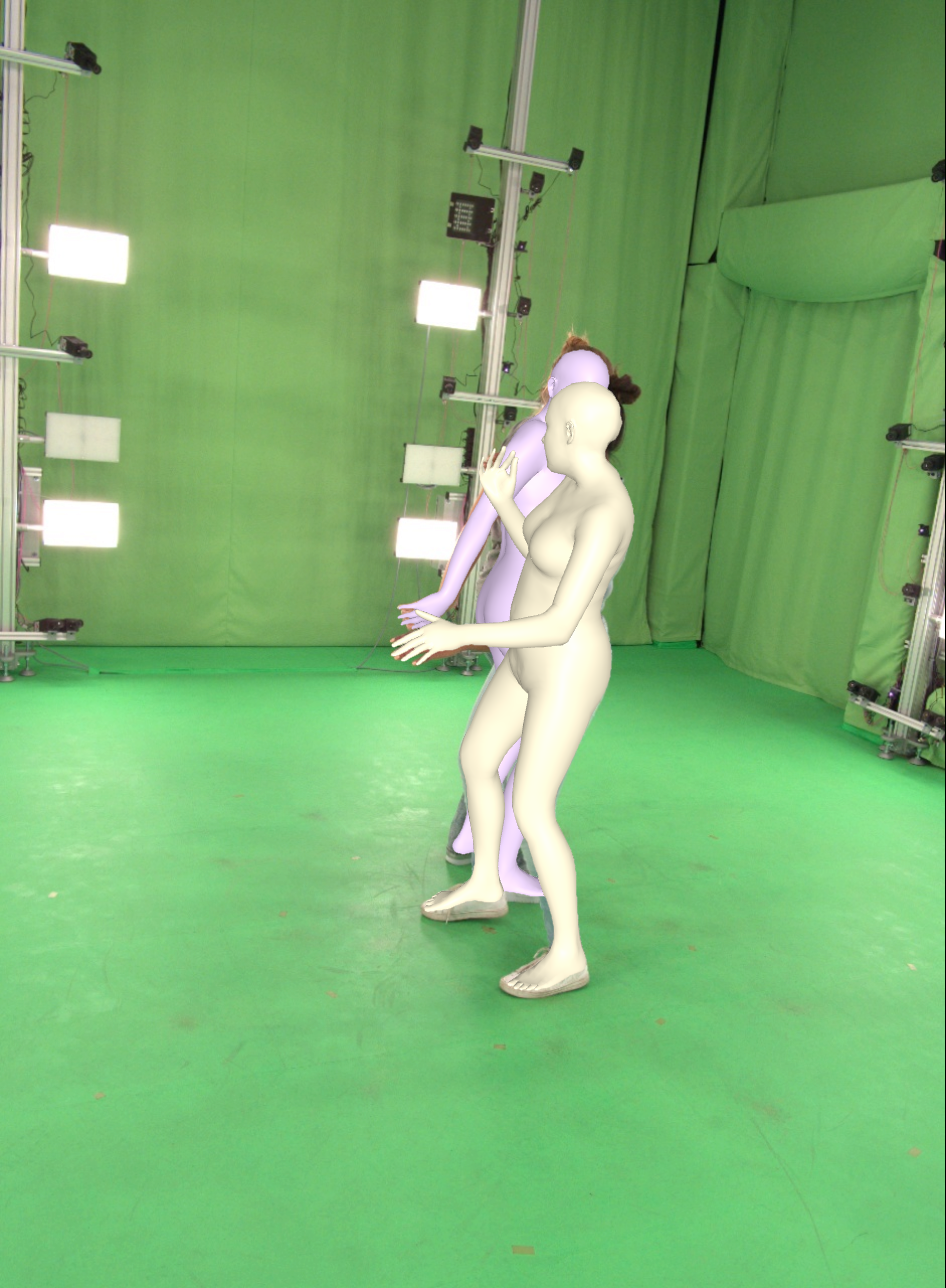} &
\includegraphics[width=0.23\textwidth]{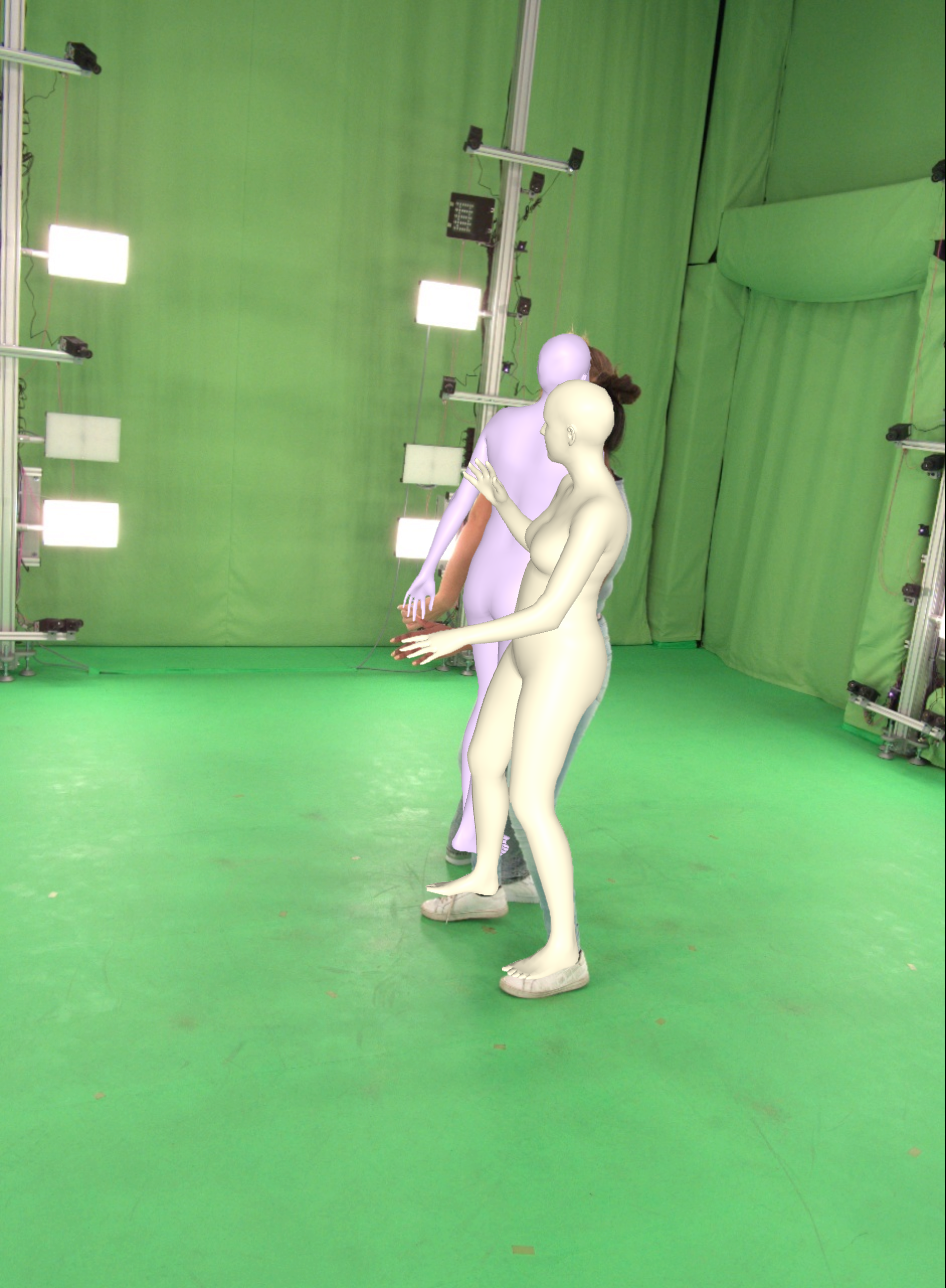} &
\includegraphics[width=0.23\textwidth]{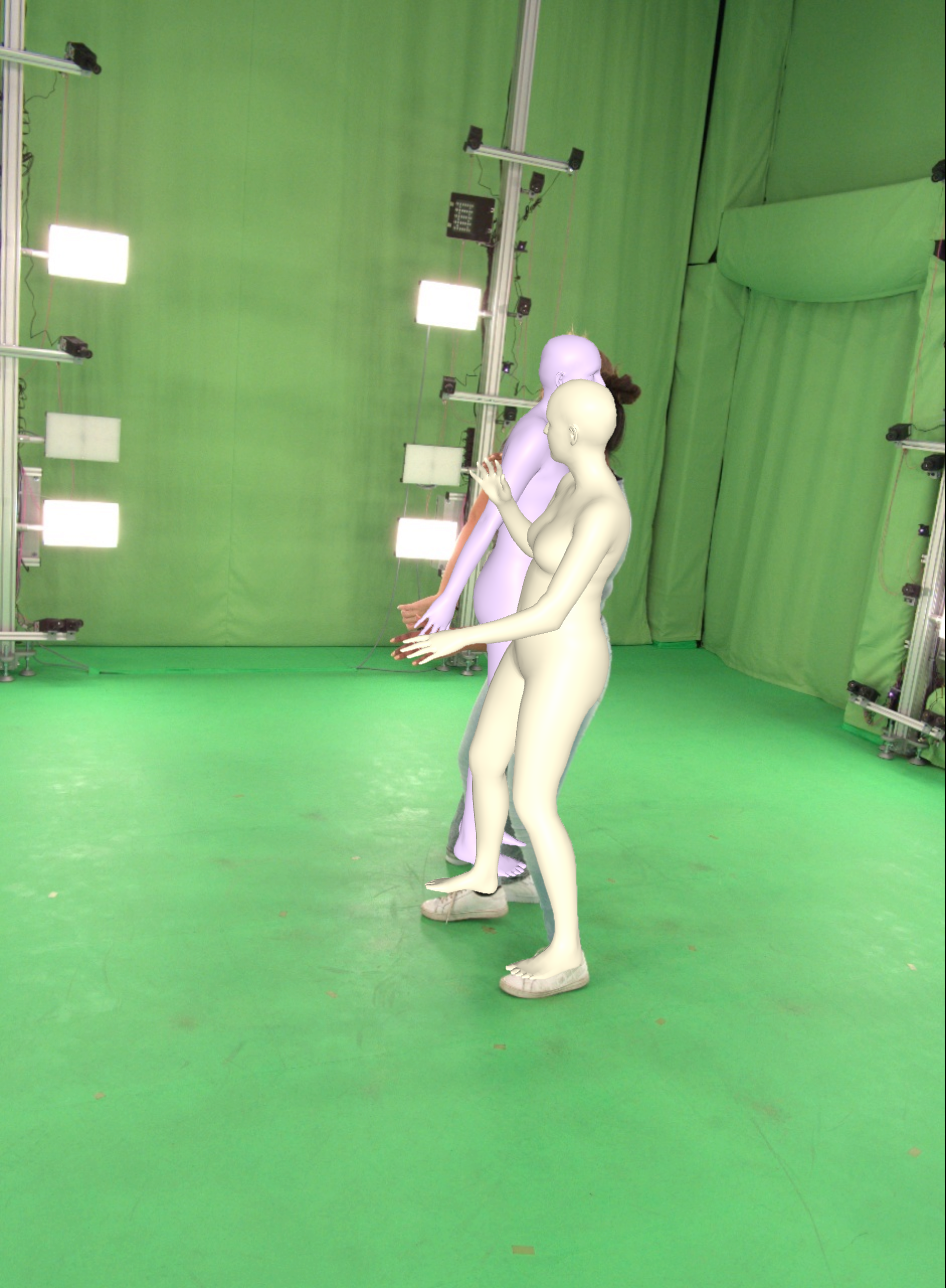} \\[4pt]
Input & GT & MA-HMR & MA-HMR+\dataname \\
\end{tabular}
\caption{\textbf{Qualitative comparison on occluded multi-person sequences.}
Each row shows one challenging frame from Hi4D.
From left to right: ground-truth mesh overlay, MA-HMR prediction, and MA-HMR+\dataname{} prediction.
}%
\label{fig:hpe_qualitative}
\end{figure*}

%% file: tables/nvs_supp.tex
\begin{table}[tb!]
\centering
\begin{tabular}{lcc}
\toprule
Method & Input Point Map & COR $\downarrow$ \\
\midrule
Ours (Kubric) & GT & 0.236 \\
Ours (\dataname) & GT & \textbf{0.124} \\
\midrule
Ours (Kubric) & DA3 & 0.928 \\
Ours (\dataname) & DA3 & \textbf{0.513} \\
\bottomrule
\end{tabular}
\caption{\textbf{Quantitative evaluation} for the multiview correspondence error on geometry-aware novel view synthesis on \dataname evaluation benchmark.}%
\label{tab:nvs_geo_cor}
\end{table}

%% file: tables/nvs_geo_master.tex
\begin{table}[tb!]
\centering
\resizebox{\linewidth}{!}{%
\setlength{\tabcolsep}{3pt} 
\begin{tabular}{@{}l cc ccc cc@{}}
\toprule
\multirow{2}*{\textbf{Method}} & \multicolumn{2}{c}{\textbf{Visual Quality}} & \multicolumn{3}{c}{\textbf{Camera Accuracy}} & \multicolumn{2}{c}{\textbf{Geometry Quality}} \\ 
\cmidrule(lr){2-3}
\cmidrule(lr){4-6}
\cmidrule(lr){7-8}
& CLIP-V~$\uparrow$ & FVD-V~$\downarrow$ & ATE~$\downarrow$ & RPE \scriptsize{trans}~$\downarrow$ & RPE \scriptsize{rot}~$\downarrow$ & CLIP-V-P~$\uparrow$ & FVD-V-P~$\downarrow$  \\ 
\midrule
Ours (Kubric) & 0.625 & 1448 & \textbf{0.202} & \textbf{0.015} & 0.207 & 0.808 & 1307 \\

Ours (\dataname) & \textbf{0.708} & \textbf{1084} & 0.231 & 0.016 & \textbf{0.164} & \textbf{0.899} & \textbf{397.4}  \\

\bottomrule
\end{tabular}
}
\caption{\textbf{Quantitative evaluation} for geometry-aware novel view synthesis on ReCamMaster evaluation benchmark.
The model trained with our \dataname performs much better in terms of visual quality and geometry quality than one trained on Kubric.
}%
\label{tab:nvs_geo_master}
\end{table}

%% file: tables/nvs_geo_master_rebuttal.tex
\begin{table}[tb!]
\centering
\resizebox{\linewidth}{!}{%
\setlength{\tabcolsep}{3pt}
\begin{tabular}{@{}ll cc ccc cc@{}}
\toprule
\multirow{2}*{\textbf{Eval Dataset}} & \multirow{2}*{\textbf{Method}} & \multicolumn{2}{c}{\textbf{Visual Quality}} & \multicolumn{3}{c}{\textbf{Camera Accuracy}} & \multicolumn{2}{c}{\textbf{Geometry Quality}} \\
\cmidrule(lr){3-4}
\cmidrule(lr){5-7}
\cmidrule(lr){8-9}
 & & CLIP-V~$\uparrow$ & FVD-V~$\downarrow$ & ATE~$\downarrow$ & RPE \scriptsize{trans}~$\downarrow$ & RPE \scriptsize{rot}~$\downarrow$ & CLIP-V-P~$\uparrow$ & FVD-V-P~$\downarrow$  \\
\midrule
\multirow{3}*{ReCamMaster} & ReCamMaster      & 0.8381 & 552.7 & 0.1417 & 0.0083 & 0.0474 & ---    & ---     \\
\cmidrule(lr){2-9}
                           & Ours (Kubric)   & 0.6404 & 1964  & 0.9424 & 0.0415 & 0.1799 & 0.9062 & 648.3 \\
                           & Ours (\dataname) & \textbf{0.6840} & \textbf{1566} & \textbf{0.4440} & \textbf{0.0020} & \textbf{0.0817} & \textbf{0.9364} & \textbf{610.2} \\
\midrule
\multirow{3}*{\dataname}   & ReCamMaster      & 0.8846 &	1698 &	0.0294 &	0.019 &	0.4128 & --- & --- \\
\cmidrule(lr){2-9}
                           & Ours (Kubric)   & 	0.7945 &	2301 &	0.0297 &	0.0135 &	0.1445 &	0.8303 &	\textbf{461.5}  \\
                           & Ours (\dataname) & \textbf{0.8108} & \textbf{2294} & \textbf{0.0203} & \textbf{0.0131} &	\textbf{0.1201} & \textbf{0.8724} & 566.5 \\
\bottomrule
\end{tabular}
}
\caption{\textbf{Quantitative evaluation for geometry-aware NVS with matched clip counts (2048) and frame number (first 49 frames).} We evaluate the same Kubric- and \dataname-trained models on the ReCamMaster and \dataname benchmarks (zero-shot for both). The \dataname-trained model still wins under the matched-budget setting.
}%
\label{tab:nvs_geo_master_rebuttal}
\end{table}